\documentclass[10pt,journal,compsoc]{IEEEtran}
\ifCLASSOPTIONcompsoc
  \usepackage[nocompress]{cite}
\else
  \usepackage{cite}
\fi
\ifCLASSINFOpdf
\else
\fi

\usepackage{times}
\usepackage{epsfig}
\usepackage{graphicx}
\usepackage{amsmath}
\usepackage{amssymb}

\usepackage{gensymb}

\usepackage{subcaption}
\usepackage{enumitem}
\usepackage{import}
\usepackage{mathtools}

\usepackage[pagebackref=true,breaklinks=true,colorlinks,bookmarks=false]{hyperref}

\usepackage{flushend}

\usepackage[dvipsnames, table]{xcolor}
\usepackage{xspace}
\usepackage{xcolor}
\usepackage{diagbox}
\usepackage{bm}
\usepackage{amssymb}
\usepackage{booktabs}

\usepackage{multirow}
\usepackage{caption}
\usepackage{float}

\definecolor{Gray}{gray}{0.9}
\definecolor{lightgray}{gray}{0.94}

\newcommand{\fbt}[1]{{\color{black}{#1}}}

\newcommand{\rbt}[1]{{\leavevmode\color{black}{#1}}}

\newcommand{\smpltransformer}{PoseBERT\xspace}
\newcommand{\poseformer}{\smpltransformer}
\newcommand{\posebert}{\smpltransformer}
\newcommand{\sspin}{MoCap-SPIN\xspace}

\newcommand{\mask}{\mu}
\newcommand{\maskseq}{\mu}

\newcommand{\inpose}{\mathbf{p}}
\newcommand{\inposeseq}{\mathcal{P}}

\newcommand{\emb}{\mathbf{x}}
\newcommand{\embmask}{\bar{\mathbf{x}}}
\newcommand{\embseq}{\mathcal{X}}
\newcommand{\PEvec}{\mathbf{\text{PE}}}

\newcommand{\outpose}{\mathbf{m}}
\newcommand{\outposeseq}{\mathcal{M}}

\newcommand{\thetaprev}{\theta^{l-1}}
\newcommand{\thetacurr}{\theta^{l}}

\newcommand{\errorgain}[1]{\color{OliveGreen}{($\downarrow$ #1)}}
\newcommand{\accgain}[1]{\color{OliveGreen}{($\uparrow$ #1)}}

\makeatletter
\DeclareRobustCommand\onedot{\futurelet\@let@token\@onedot}
\def\@onedot{\ifx\@let@token.\else.\null\fi\xspace}

\def\eg{\emph{e.g}\onedot} 
\def\ie{\emph{i.e}\onedot}

\def\etal{\emph{et al}\onedot}
\makeatother

\renewcommand{\paragraph}[1]{\vspace{0.02cm}\noindent\textbf{#1}}

\begin{document}
%
\title{PoseBERT: A Generic \fbt{Transformer} Module  for \\ Temporal 3D Human Modeling}
%
%
%
%

\author{Fabien Baradel,
        Romain Br\'egier,
        Thibault Groueix,
        Philippe Weinzaepfel, \\
        Yannis Kalantidis,
        Gr\'egory Rogez
\IEEEcompsocitemizethanks{
\IEEEcompsocthanksitem NAVER Labs Europe\newline
6 chemin de Maupertuis, 38240 Meylan, France\newline
E-mail: firstname.lastname@naverlabs.com
}
}

\markboth{IEEE TRANSACTIONS ON PATTERN ANALYSIS AND MACHINE INTELLIGENCE, 2022}%
{Baradel \MakeLowercase{\textit{et al.}}: PoseBERT}
%



\IEEEtitleabstractindextext{%

\begin{abstract}
\noindent
Training state-of-the-art models for human pose estimation in videos requires datasets with annotations that are really hard and expensive to obtain.
Although transformers have been recently utilized for body pose sequence modeling, related methods rely on pseudo-ground truth to augment the currently limited training data available for learning such models.
In this paper, we introduce \posebert, a transformer module that is fully trained on 3D Motion Capture (MoCap) data via masked modeling.
It is simple, generic and versatile, as it can be plugged on top of any image-based model to transform it in a video-based model leveraging temporal information. We showcase variants of \posebert with different inputs varying from 3D skeleton keypoints to rotations of a 3D parametric model for either the full body (SMPL) or just the hands (MANO).
Since \posebert training is task agnostic, the model can be applied to several tasks such as pose refinement, future pose prediction or motion completion  \emph{without finetuning}. Our experimental results validate that adding \posebert on top of various state-of-the-art pose estimation methods consistently improves their performances, while its low computational cost allows us to use it in a real-time demo for smoothly animating a robotic hand via a webcam.
Test code and models are available at \url{https://github.com/naver/posebert}.
\end{abstract}


\setcounter{figure}{0}
\begin{center}
\centering\small
\includegraphics[width=.98 \linewidth]{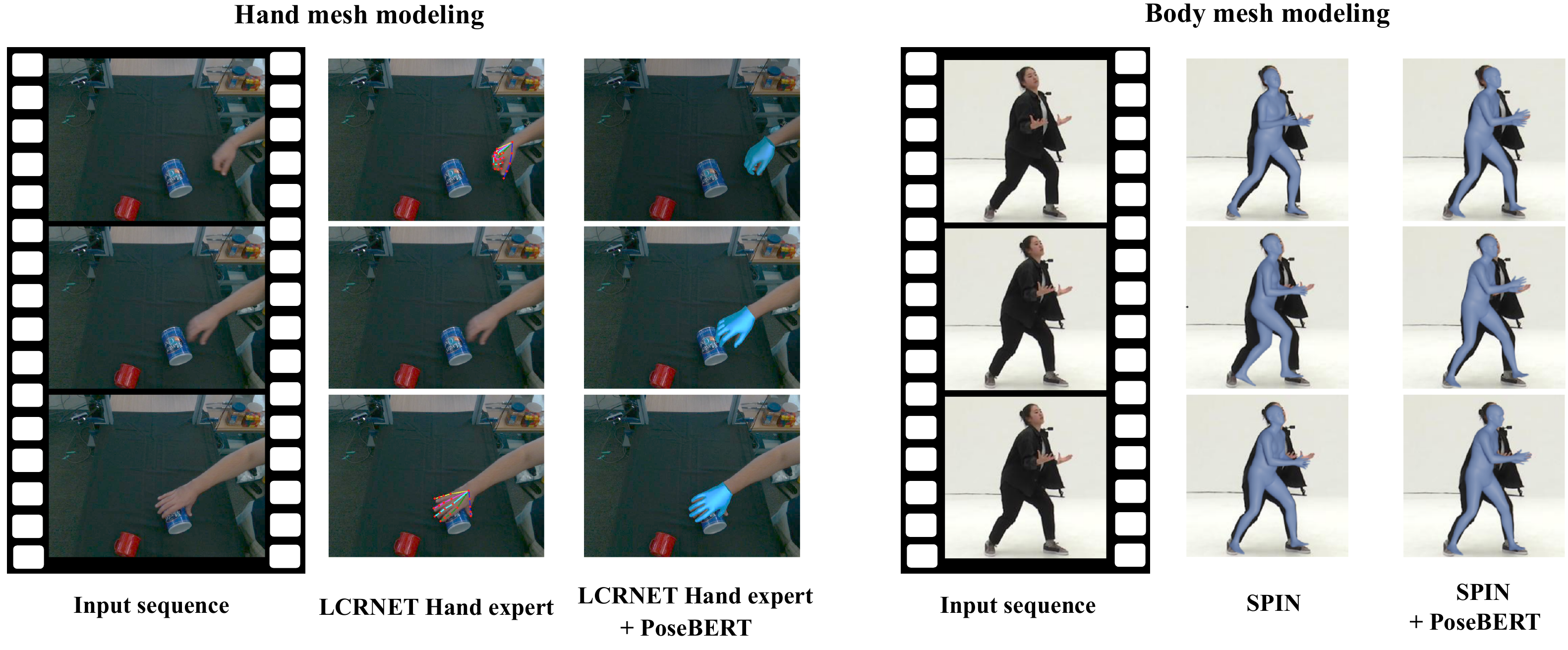}
\captionof{figure}{\label{fig:teaser}
\textbf{
Qualitative comparison between image-based pose estimators and the proposed \posebert model.}
We show qualitative results on hand mesh recovery (left) and human body mesh recovery (right) when plugging \posebert ontop of LCR-Net~\cite{lcrnet++} hand expert (used in~\cite{dope,hand_challenge2019}) and SPIN~\cite{spin} respectively.
\posebert allows to fill-in missing detections (left, second row) and refine the estimated poses to better align with the input images (right).
}

\end{center}
}

\maketitle

\IEEEdisplaynontitleabstractindextext

%
\IEEEpeerreviewmaketitle

\IEEEraisesectionheading{\section{Introduction}\label{sec:introduction}}

\fbt{
\IEEEPARstart{S}{tate}-of-the-art methods that estimate 3D body pose and shape given an image~\cite{hmr,nbf,spin,pose2mesh,Li_2021_CVPR,lin2021mesh} or a video~\cite{vibe,meva,li2021task} have recently shown impressive results.
}
A major challenge when training models for in-the-wild human pose estimation is  data: collecting large sets of training images with ground-truth 3D annotations is cumbersome as it requires setting up IMUs~\cite{3dpw}, calibrating a multi-camera system~\cite{huang2017} or considering static poses~\cite{mc3dv}.  
Only 2D information -- such as the location of 2D keypoints or semantic part segmentation -- can reasonably be manually annotated on in-the-wild data.
\fbt{Current 3D pose estimation methods leverage such 2D annotations by training their model to minimize a 2D reprojection error~\cite{hmr}, by generating 3D pseudo-labels with an optimization-based method~\cite{smplify} beforehand and curating the obtained ground-truth~\cite{unite}, or by running the optimization inside the training loop~\cite{spin,rempe2021humor}.
}
This lack of real-world data with 3D annotations is even more critical for videos, making difficult the use of recent temporal models such as transformers~\cite{transformer} which are known to require large datasets for training.
Current approaches for video-based human pose estimation~\cite{vibe, tcmr, hmmr}  rely on weak  (2D pose) and/or pseudo 3D ground-truth annotations on rather small video datasets.

At the same time, Motion Capture (MoCap) -- widely used in the video-game and film industry -- offers a solution to create large corpus of motion sequences with accurate ground-truth 3D poses. Recently, several of these MoCap datasets have been unified into the large AMASS dataset~\cite{amass} using SMPL~\cite{smpl} -- a  differentiable  parametric human mesh model used in numerous state-of-the-art human mesh recovery methods~\cite{smplify,hmr,nbf,spin,pose2mesh}.
The use of large-scale MoCap data for video-based human pose estimation has been mainly focused on  
improving the realism of the sequences of estimated 3D poses~\cite{hmmr,vibe}. In this paper,  we exploit them for learning better temporal models and  introduce \posebert, a transformer module that  is purely trained on 3D MoCap data.  We learn the parameters of \posebert using masked modeling, similar to BERT~\cite{bert}, and end up with a generic and highly versatile module that can be used without finetuning for a number of tasks and datasets, and for full body or just hand pose sequence modeling. In particular, \posebert can be plugged on top of any state-of-the-art image-based pose estimation model in order to transform it in a video-based model leveraging temporal information.
Figure~\ref{fig:teaser} (left) shows a qualitative comparison between hand keypoints detected by LCR-Net~\cite{lcrnet++} and the output of \posebert which allows to fill-in the missing detections and transform the keypoints into hand meshes. Figure~\ref{fig:teaser} (right) depicts a qualitative comparison between an image-based human mesh recovery method (SPIN~\cite{spin}) and the impact of \posebert when applied on top of it, which allows to smooth and refine the predictions by leveraging temporal information.
Although a specific instantiation of the \posebert module for SMPL inputs was originally introduced among other contributions in our recently published work of~\cite{leveraging_mocap}, in this journal paper we widely extend the formulation, analysis and evaluation of the module. We first define a generalized version of \posebert that works on a variety of input and output signals beyond SMPL parameters, \ie, hand meshes as well as 3D skeleton keypoints. We propose to also estimate the 3D pose from a camera-centric point of view by predicting the 3D location of the person in the scene, and we end up with a generic, task-independent module that can solve a range of downstream tasks without fine-tuning, such as denoising pose sequences, recovering missing poses, refining an initial pose sequence in 3D, 
motion completion or future pose prediction. We further show how \posebert can be used in a plug-and-play fashion on top of different pose \fbt{estimation} methods and works for both human body and hand modeling. 
Finally, going beyond ground-truth person bounding boxes that are typically used in most of the human mesh recovery state-of-the-art methods including~\cite{leveraging_mocap}, here, we also evaluate a more realistic scenario where both person detection boxes and pose regressions are extracted in an automatic way using the multi-person LCR-Net\cite{lcrnet++}.
\fbt{We also experimented with the recent HybrIK~\cite{Li_2021_CVPR}, first detecting the persons using Faster-RCNN~\cite{girshick2015fast} detector as in~\cite{ochmr} and then estimating 3D human poses individually.
}
We show that in this scenario where predictions are very noisy and many detections are in practice missing, \posebert leads to even higher gains.

\subsection{Contributions}
\label{sec:contributions}

In summary, our contributions are the following:
\begin{itemize} 
  \item We introduce \posebert, a transformer-based module for pose sequence modeling of monocular RGB videos, that can be trained without the need for any cumbersome RGB image or frame pose annotations. Instead, we \textbf{leverage MoCap data for training}, a modality that is in comparison relatively easy to acquire and large scale MoCap datasets are already available.
  Our method works for both human body and hand modeling; in fact, it can be used for any usecase where there exists a 3D parametric model and MoCap data are available to train it.
  
  \item We learn \posebert parameters with masked modeling and end up with a generic, \textbf{task-independent model} that can be used out-of-the-box, \ie, \textit{without fine-tuning} on a number of downstream tasks such as denoising pose sequences, recovering missing poses in a sequence, refining an initial pose sequence in 3D, motion completion or future frame prediction. \posebert is plug-and-play, independent of the frame-based method used to extract input poses and can trivially handle frames with missing predictions. 
  
  \item We extensively evaluate a number of variants of \posebert with different input types varying from 3D skeleton keypoints to rotations of a 3D parametric model for body (SMPL) or hands (MANO), on a large number of downstream tasks and datasets. Some highlights are that a) \posebert always improves the performance on pose refinement whatever the off-the-shell image-based method taken as input with improvements ranging from 1.0 to 10.3 points in PA-MPJPE; b) \posebert brings a relative gain of 10\% to 50\% for the task of future pose prediction compared to strong baseline, for future horizons ranging from 5 to 30 frames. \textit{Our method can predict plausible future pose up to 1 second in the future.}

  \item We present extensive ablations for the proposed module and its training strategy as well as a number of analysis, including a study on missing frames and motion completion. 
  
  \item \posebert has a low computational cost which allows us to use this temporal model in an online manner in real-time (30 fps), while a forward pass takes approximately 5ms.
  Adding \posebert on a top of a standard image-based method adds a 10\% computational overhead in term of FLOP while bringing a robust motion recovery.
  
  The robustness and the low computation cost of the module has further allowed us to build a real-world application that utilizes \posebert for \textbf{robotic teleoperation}. In that, the hand pose is estimated from the monocular RGB video stream of a webcam, and then transfered in real time to a robotic hand gripper.
\end{itemize}
\section{Related work}
\label{sec:related}

In this section, we briefly review related works on the estimation of human parametric models in videos, the task of pose sequence generation, and the use of MoCap data for pose estimation.

\paragraph{3D human pose estimation in videos with parametric models.}
\fbt{
3D human pose estimation has been mainly studied from a single RGB image point of view \cite{spin,Li_2021_CVPR,lin2021mesh,song2020human,akhter2015pose,Yuan_2022_CVPR,Choi_2022_CVPR,lin2021end,mehta2020xnect}. However, many uncertainties in 3D pose due to depth ambiguities~\cite{Kundu_2022_CVPR} or strong occlusions~\cite{ochmr} can be resolved by taking into consideration neighboring frames in a video.
Recent works~\cite{dabral2018learning,li2019boosting,khurana2021detecting,Liu_2022_CVPR,choi2021beyond} in video-based human pose estimation therefore aim at solving this issue by leveraging temporal information.
}
Most current approaches for video-based human 3D pose estimation rely on pose estimates~\cite{Huang17, videokinetics,Jiang_2021_CVPR}  or pose features~\cite{vibe,sun2019human,meva, hmmr,tcmr} derived from each frame independently. Their predictions are conditioned on these sequential input data. Arnab \etal~\cite{videokinetics} proposed an optimization-based strategy to handle human pose estimation in videos.
In HMMR~\cite{hmmr}, features from consecutive frames are fed to a 1D temporal convolution, while VIBE~\cite{vibe} uses recurrent neural network, namely Gated Recurrent Unit (GRU), together with a discriminator at the sequence level.
The network is trained on different in-the-wild videos and losses are similar to the ones employed for images and previously described, \ie, mainly applied on keypoints.
A similar architecture with GRU is used in TCMR~\cite{tcmr}, except that 3 independent GRUs are used and concatenated, one in each direction and one bi-directional in order to better leverage temporal information.
MEVA~\cite{meva} estimates motion from videos by also extracting temporal features using GRUs and then estimates the overall coarse motion inside the video with Variational Motion Estimator (VME).
For hand pose, ~\cite{seqhand} use LSTMs on top of image features to predict MANO parameters; Their model is pretrained on synthetic data before being finetuned on real data.
Recently, Pavlakos \etal~\cite{thmmr} have proposed to use a transformer architecture~\cite{transformer}.
To obtain training data, \ie, in-the-wild videos annotated with 3D mesh information, they use the smoothness of the SMPL parameters over consecutive frames to obtain pseudo-ground-truth. In terms of architecture, the transformer is used to leverage temporal information by modifying the features.
We also consider a transformer architecture, but we apply it directly to the pose predictions of an image-based model. This has the great advantage of being directly trainable on MoCap data (e.g. on AMASS for body pose) and pluggable on top of any image-based method.
%
Related to our work, Jiang \etal~\cite{Jiang_2021_CVPR} also consider a transformer network trained with masked language modeling.
They train their network using pseudo ground-truth 3D annotations, obtained using a 3D uplift process from 2D poses estimated on RGB videos.
A disadvantage of such approach is that it does not guarantee the plausibility of the pseudo ground-truth, contrary to MoCap data.

\paragraph{Pose sequence generation.}
\fbt{
Although it does not specifically target the task of pose sequence generation, our \posebert transformer architecture can also be used to generate upcoming poses in the near future. For pose sequence generation, previous methods either condition on the beginning of a sequence of pose~\cite{Aliakbarian_2021_ICCV,Mangalam_2021_ICCV,wei2019motion,gopalakrishnan2019neural} or on some predefined labels like human actions~\cite{petrovich2021action,aksan2019structured,aksan2021spatio}.
Most current approaches~\cite{hpgan,hernandez2019human,habibie2017recurrent,mojo} are based on GAN~\cite{GAN} or Variational Auto-Encoder (VAE)~\cite{VAE}.
More recently, a cross-modal transformer architecture has been proposed in~\cite{aistpp} to generate human pose sequences conditioned on music.
Cao \etal \cite{cao2020long} also propose to take the scene context into account for predicting long-term human motion.
In our case, \posebert being deterministic, it cannot handle long-term future prediction.
However, we show that it is possible to apply it to predict near future or complete missing frames, without any particular conditioning.
}

\paragraph{Pose sequence completion}
\fbt{
 is a well-studied field \cite{harvey2020robust,kaufmann2020convolutional} where the task corresponds to fill in-betweening of a pose sequence where only few frames of a sequence are known.
In a concurrent work, Duan \etal~\cite{duan2021single} propose a transformer architecture to perform pose sequence completion between the first and the last frames of a sequence using a position and angle representation per keypoint.
}
In our case, \posebert  does not restrict to completion and uses a masked pose sequence modeling similar to BERT for text, with additional Gaussian noise, to better generalize to more tasks such as denoising or generation.


\paragraph{Use of MoCap data.}
\fbt{
3D human modeling with synthetic data is a well-explored idea and many papers proposed to apply the learned models to solve various 3D human related tasks (e.g., 3D pose estimation, human mesh reconstruction) \cite{surreal,bodynet,STRAPS2020BMVC,Xu_2019_ICCV,sim2real}.
MoCap data can be used to generate diverse synthetic images with ground-truth annotations using a rendering engine -- as in SURREAL~\cite{surreal} or ~\cite{rogez2018image}.
The inherent domain shift between synthetic and real-world images has often limited  the use of such data to pretraining~\cite{bodynet} or finetuning~\cite{leveraging_mocap}. To avoid the sim2real problems, some methods have proposed to use proxy representations about the person’s appearance, e.g. IUV maps in~\cite{Xu_2019_ICCV} or silhouette and 2D keypoints in~\cite{STRAPS2020BMVC}, or the motion, e.g. optical flow and 2D keypoint movements in ~\cite{sim2real}.

In~\cite{lcrnet++},  pseudo-groundtruth 3D annotations are obtained by matching 2D annotations with reprojected 3D MoCap data.
Similarly, the learning-by-synthesis approach in \cite{xu2021monocular} learns a 2D-3D mapping function to lift 2D poses into 3D using MoCap data and random  2D projections. 

MoCap data have also been used to augment the realism of human pose estimates.
For example, Kanazawa \etal and Kocabas \etal used MoCap data to train a pose discriminator for image-based~\cite{hmr} or video-based~\cite{vibe} models.
Such discriminator enforces the model to predict realistic outputs, but it does not improve the diversity of predicted poses.

Recent works aim at leveraging MoCap data to learn a human motion priors such that they do not rely on images or videos renderings and thus mitigates the domain shift issue \cite{rempe2021humor,li2021task}.
Rempe \etal \cite{rempe2021humor} propose a robust approach. However the proposed framework rely on an optimization and is therefore very slow. We also exploit MoCap data to train \posebert but our framework is extremely fast and runs in real time.

}

\section{PoseBERT}
\label{sec:method}

\begin{figure*}[!th]
\centering
\includegraphics[width=.9 \linewidth]{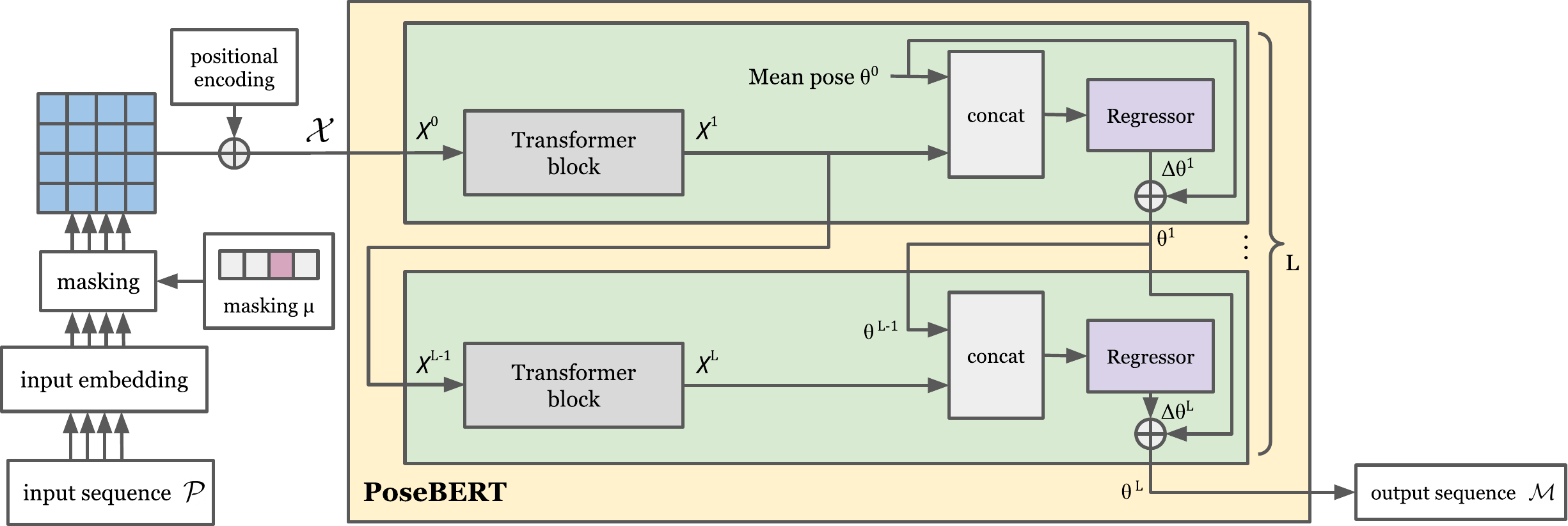} 
\caption{\label{fig:transformer}\textbf{The \posebert architecture}. 
The input is a representation of a temporal sequence of $T$ poses (\eg keypoints or pose parameters of a parametric model), while the output we consider is the pose parameters of a parametric model along the same sequence of $T$ frames.
\smpltransformer basic block is repeated $L$ times. The regressor parameters are shared across the $T$ inputs and the $L$ blocks.
We regress the pose starting from the mean pose of the parametric 3D model.
For a sake of clarity we do not show the translation branch.
}
\end{figure*}

In this section we present \posebert, a transformer-based module that is able to transform a noisy and/or incomplete sequence of poses into a smooth and coherent sequence of meshes. We first present architectural details of the module in Section~\ref{sec:posebert_architecture} and the proposed training framework via masked modeling and/or denoising the input sequence in Section~\ref{sec:posebert_training}. We then showcase a number of variants of \posebert with different input types varying from 3D skeleton keypoints to rotations of a 3D parametric model for body (SMPL) or hands (MANO) in Section~\ref{sec:posebert_io}.




\subsection{Overview}
\label{sec:posebert_overview}

The \posebert module takes as input a sequence of poses $\mathcal{\inposeseq} = \{ \inpose_1, \ldots, \inpose_T \}$, where $T$ denotes the sequence length. Without loss of generality, we assume that the sequence corresponds to $T$ frames from a video, and that each pose is extracted via an off-the-shell, image-based pose estimator that operates on each frame independently. We further assume each pose to be represented by a high-dimensional vector; the exact nature of the input representation is not restricted and can for example correspond to any parametric pose model for either the whole body or for just the hand (see Section~\ref{sec:posebert_io}).

We argue that such image-based pose estimators highly suffer in the presence of motion blur or occlusions; some poses can therefore be highly noisy or simply missing. Our goal is to learn a module that takes such a noisy input pose sequence $\inposeseq$ and outputs a more temporally coherent output sequence $\outposeseq$, inferring any missing predictions if needed. Let the sequence of output meshes $\outposeseq = \{ \outpose_1, \ldots, \outpose_T \}$, the \posebert module implements the following mapping:
\begin{equation}
    \outposeseq = \text{\posebert} (\inposeseq).
    \label{eq:posebert_generic}
\end{equation}
Similar to the input representations, the nature of the output representations may also vary; in this paper, we explore output sequences defined on 3D mesh for hand using the parametric model MANO~\cite{romero2017mano} or body using the parametric model SMPL~\cite{smpl}.

\paragraph{Masking and modeling of missing frames.} The learning of the \posebert parameters is based on masked modeling of parts of the input sequence. Although masking is controlled and simulated during training, it may also exist during testing in the case of missing predictions.
Let $\maskseq = \{ \mask_1, \ldots, \mask_T \}$ denote a binary vector indicating if a pose is available or missing for each timestep; $\mask_t=1$ indicates that  a pose $\inpose_t$ is provided by the image-based estimator, while if $\mask_t=0$, the pose for timestep $t$ is missing. For training as well as for testing in the presence of missing predictions, Equation~(\ref{eq:posebert_generic}) becomes:
\begin{equation}
    \outposeseq = \text{\posebert} (\inposeseq, \maskseq).
    \label{eq:posebert_generic_with_mask}
\end{equation}

\subsection{Architecture of the \posebert module}
\label{sec:posebert_architecture}

In this section, we present the architecture of \posebert, a transformer-based~\cite{transformer} model composed of $L$ blocks. An overview of the architecure is shown in Figure~\ref{fig:transformer}. First, the input sequence is projected to a sequence of inputs embeddings, while a positional encoding is added. For the first layer of \posebert, we set all three inputs of the transformer (query, key and value) to the sequence of input embeddings. Then, each subsequent layer takes as input the output of the previous layer. The output of the transformer block is concatenated with the pose estimation from the previous layer and fed to a regressor that updates the current estimate for the pose. Below we present the main components in detail.

\paragraph{Input embedding.}
We first embed each pose $\inpose \in \inposeseq$ in a $D$-dimensional space using a linear projection. If a pose is missing, we replace the embedding by a special learnable token denoted $\embmask$. Similarly to~\cite{transformer}, we also add a learnable 1-D positional encoding $\PEvec$ to the input of the first layer.
Specifically, the sequence of input embeddings $\embseq = \{ \emb_1, \ldots, \emb_T \}$ is given by: 

\begin{equation}
    \emb_t = 
\begin{dcases}
    \embmask + \PEvec_t,& \text{if } \mask_t=0 \\
    e(\inpose_t) + \PEvec_t, & \text{otherwise}
\end{dcases}
\end{equation}
for all $t = 1, .. T$, where $e(\inpose_t) = W_{e} \inpose_t + \mathbf{b}_{e}$ is the learnable linear projection of input pose $\inpose_t$.

\paragraph{Modeling temporal context with a transformer block.}
The sequence of pose embeddings $\embseq$ is iteratively updated with contextual information using a series of vanilla transformer blocks \cite{transformer}. Each transformer block is composed of a multi-head scaled dot-product attention mechanism and a feed-forward module.
The output $\embseq^l = \{ \emb_1^l, \ldots, \emb_T^l \}$  of the $l^{th}$ transformer block (with $\emb_t^l \in \mathrm{R}^D$) is given by:
\begin{equation}
    \embseq^{l} = \text{Transformer}^{l}(\embseq^{l-1}).
\end{equation}
with $\embseq^0 = \embseq$ by convention.
Similar to the original transformer~\cite{transformer}, we use layer normalization before self-attention modules; all other design choices also follow the original transformer architecture. 

\paragraph{Iterative pose regression.}
Estimation of the 3D pose parameters $\theta$ is done independently for each timestep and proceeds in an iterative way.
Specifically, let $\theta_t = \theta_t^L$ denote the final pose estimation for timestep $t$, \ie at the final layer $L$ of \posebert. At each layer $l = 1, .., L$ and given the estimation of the previous layer $\thetaprev_t$, the regressor module updates the pose estimation using the following mapping:
\begin{gather}
    \Delta \thetacurr_t =  \text{Regressor}([\emb_t^l, \thetaprev_t]) \\
    \thetacurr_t = \thetaprev_t + \Delta \thetacurr_t \\
    \theta_t^0 = \theta_\text{mean}
\end{gather}
where $[\cdot,\cdot]$ denotes concatenation, $\thetacurr$ denotes the pose parameter estimation after layer $l$ and $\theta_\text{mean}$ denotes the mean pose.
\fbt{
For the $\text{Regressor}(\cdot)$ function we use a Multilayer Perceptron (MLP) with the same architecture as in~\cite{hmr}.
In our case, however, the regressor at each layer of \posebert is \textit{not iterative}, but a simple feedforward network.
We instead ``unroll'' the regressor iterations throughout the $L$ layers of \posebert, effectively performing $L$ ``iterations'' of the regressor overall.
We show in the experiments section that this strategy reaches better performance compared to adding the regressor at the end of the network as in \cite{spin,hmr,Pavlakos_2022_CVPR}.
Similar to \cite{hmr} we inject some inductive bias into our regression framework by initializing the pose estimation $\theta_0$ with the mean pose $\theta_\text{mean}$.
}
Regressor parameters are shared across all $T$ inputs.

In Section~\ref{sec:experiments}, we ablate a number of parameters of this design. First, and similar to~\cite{hmr}, the regressor of \posebert could be applied iteratively  at each layer. We found however that a single iteration performs equally well. Moreover, we experiment with versions of \smpltransformer where the regressor parameters are shared across the $L$ blocks, thus reducing the number of learnable parameters.

\paragraph{Iterative translation regression.}
While in this work we are mainly interested in estimating the pose of a human in a body-centric reference frame, we also provide an estimation of its 3D location relative to the camera.
Regressing the location in Euclidean space is not a straightfoward operation thus we choose to predict the 2D location $(x_{t,2d},y_{t,2d})$ of the root of the parametric model in the image, and its \textit{nearness} $n_t=log(1/z_t)$ to the camera (where $z_t$ is its distance to the camera plane).
We regress these parameters $(x_{t,2d},y_{t,2d},n_t)$ in the same way as the other pose parameters.
\fbt{
The final location $\gamma_t=(x_t,y_t,z_t)$ expressed in camera coordinates system is obtained using the inverse camera projection mapping, derived from the camera intrinsic parameters (focal length and principal point).
We assume that these camera intrinsic parameters are fixed such that the focal length is equal to 1500 mm and the principal point is the center of the image.
}

\subsection{Learning the parameters of \posebert }
\label{sec:posebert_training}

\begin{figure*}[!ht]
\centering
\includegraphics[width=.8\linewidth]{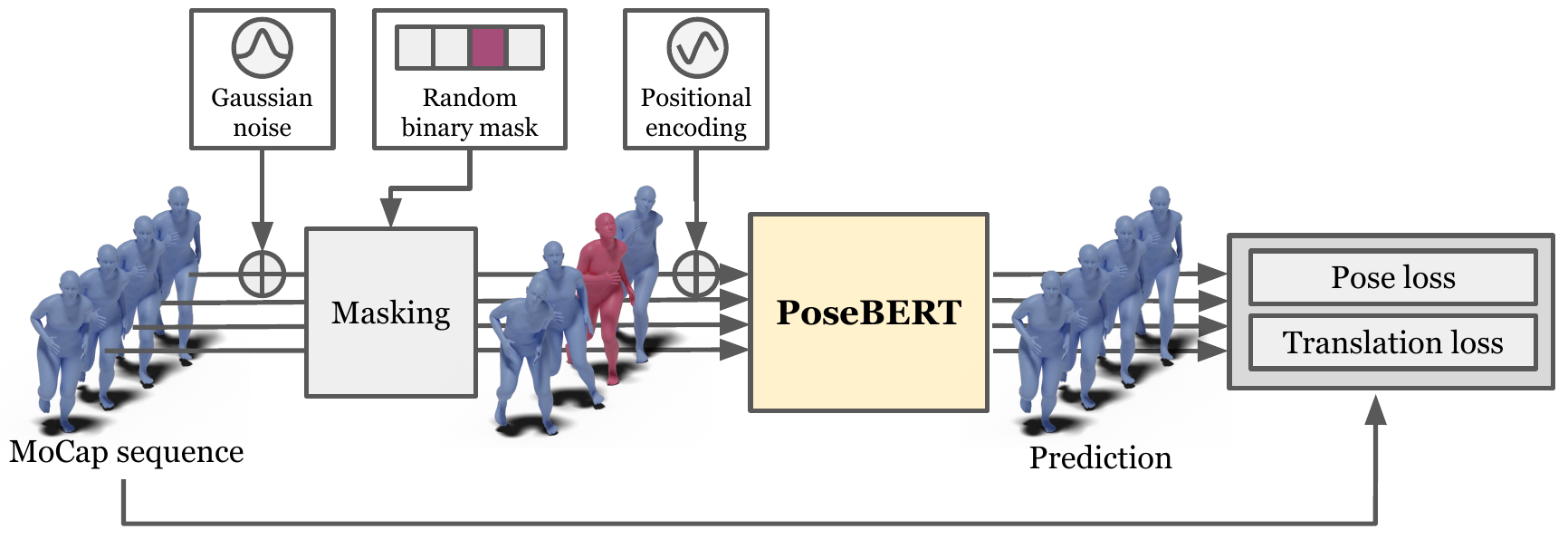} \\
\caption{\label{fig:temporal}\textbf{Learning \smpltransformer with masked modeling.} To allow the model to learn temporal dynamics and similar to~\cite{bert} we \emph{mask} part of the input and either replace it with a learnable masking token or a random pose. The mask is a random $T$-dimensional binary vector that specifies which timestamps will be masked.
}
\end{figure*}


We employ a training objective which can be split into two parts: 1) inferring missing poses and 2) denoising existing poses.
For solving these two tasks, the network should contextualize each input embedding.
We create pretext tasks to learn these skills, by masking parts of the poses and by introducing artificial noise in MoCap pose sequences. While simple, our noise model shares some similarity with the errors made by image-based methods.
as validated by our experiments.


\paragraph{Masked modeling task. }
Looking at context can help to correct errors of models based on single image frame.
By inputting a sequence of poses to \poseformer,  we want to be able to learn \emph{temporal} dynamics.
To do so, we utilize the masked modeling task, one of the self-supervised tasks that the now ubiquitous BERT~\cite{bert} model uses to learn language models.
Part of the input is masked before it is fed to the model.
The correct output is expected to be recovered using the rest of the sequence.
In practice, we randomly sample the percentage of timesteps to mask between a lower and a upper bound.
We also employ a per-block masking and found it quite important for producing robust and smooth pose sequences.
This allows to make sure that the network is learning to interpolate and not only copy-pasting the pose from the last observed timestep.
We ablate these masking strategy in Section \ref{sec:experiments}.


\paragraph{Denoising. }
Even if the image-based model estimation is providing a pose for a timestep, the pose can be approximate or not plausible due to \eg occlusions or motion blur.
To increase the robustness of \posebert to such errors, we inject noise to the input pose $\inposeseq$.
A first solution that we find consists in injecting Gaussian noise to the input pose.
Second, we also replace a percentage of the poses by picking random poses from a different sequence belonging to the same batch.
This simulates real situations such as occlusions in multi-person scenes or when two hands are interacting or close to each other.
We ablate these two noise injection strategies in Section \ref{sec:experiments}.

\paragraph{Loss. }
Figure~\ref{fig:temporal} shows an overview of the training process.
The inputs are first masked and noise is added.
They are then processed by \posebert.
\posebert outputs are then compared to the original clean inputs and -- similarly to image-based models -- we define a training loss function based on the reconstruction error of the pose and translation parameters
for every timestep.
Overall, the reconstruction loss $L$ used to train \posebert is
\begin{equation}
\mathcal{L} = \mathcal{L}_{pose} + \mathcal{L}_{translation},
\end{equation}
with
\begin{gather}
    \mathcal{L}_{pose} =  \sum_{t=1 \dots T} || \theta_t -  \theta_t^{'}  ||^2,\\
    \mathcal{L}_{translation} =  \sum_{t=1 \dots T} || \gamma_t -  \gamma_t^{'}  ||^2, 
\end{gather}
and where $\theta_t^{'}$ and $\gamma_t^{'}$ correspond to the pose and translation ground-truth.

The pose $\theta$ is expressed as a set of rotation matrices obtained after running the Gram-Schmidt orthonormalization on the 6D vector representation output \cite{rotation, bregier2021deepregression}.




\subsection{\posebert for keypoints, meshes and hands}
\label{sec:posebert_io}

\subsubsection{Input representations}
\paragraph{Parametric 3D models.}
\label{sec:parametric_3d_models}
In this paper, we predict poses using a differentiable parametric 3D model which -- given a pose parameter $\theta$ and a shape parameter $\beta$ -- can output a triangulated 3D mesh $M(\theta,\beta)$.
In this study, we focus only on the pose $\theta$ and keep the shape $\beta$ constant.
The output of \posebert are thus pose parameters $\theta_{1 \dots T}$ which are then map to meshes.

For human body we use the SMPL model \cite{smpl}, and for human hand we employ the MANO model \cite{romero2017mano}.
Our method is generic and can be adapted to predict the parameters of any parametric model for pose~\cite{romero2017mano,smpl,SMPL-X:2019,STAR:2020}.

\paragraph{Input 3D representation. }
We use different representations for the input poses of \posebert, depending on their modalities.
If these poses are expressed using a parametric model, we provide as input the set of 3D orientations parameters, represented as 6D vectors~\cite{rotation}.
The representation for a pose is in this case of size $6K$ ($K=23$ for SMPL, $21$ for MANO).
If the input pose consists of a set of 3D keypoints locations, we first normalize the 3D pose such that each bone length is equal to the mean bone length.
We follow the skeleton tree traversal for performing this bone length normalization starting from the root joint (the hip for the body and the wrist for the hand).
Second, we compute the 3D rotation aligning -- for the human body -- the spine with the Y axis and the shoulders with the X axis similar to \cite{2sagcn2019cvpr}.
For the hand, we compute the rotation aligning  the middle finger with the Y axis and the basis of the index and the pinky with the X axis.
We apply this transformation to the set of normalized keypoints, and use them as input for \posebert.
We concatenate this input with a 6D representation of the estimated rigid transformation, such that we get a final representation of size $3K+6$, where $K$ is the number of input keypoints.

\paragraph{Input 2D representation. }
We are also interested in estimating the 3D pose from a camera-centric point of view by predicting the 3D location of the person/hand in the scene.
However, the input 3D representation described above is body-centric and thus it does not provide any cues about where the person/hand is located in the scene.
We propose to also input the estimated 2D poses to \posebert as additional source of information.
The intuition is that if a person/hand is far away from the camera then its 2D pose should located in a small part of the frame.
On the opposite if the person/hand is close to the camera its 2D pose should be expressed in a large region of the frame.
\fbt{
During the training procedure, we obtain the 2D pose by projecting the 3D pose into the camera plane by assuming camera where the intrinsic/extrinsic parameters are fixed during the training and testing stage.
The camera is located at $(0,0,0)$ in the world coordinates system and its rotation in rotation matrix representation is the identity matrix.
We assume a focal length equals to 1500 mm and the principal point is the center of the frame.
We normalize the 2D pose between -1 and 1 to be invariant to the image size such that the center of the frame corresponds to (0,0).
}
The 2D input representation is the concatenation of the normalized 2D pose and is of size $2K$ where $K$ is the number of joints.
Finally the overall input representation is the concatenation of the 2D and the 3D representations.


\section{Experiments}
\label{sec:experiments}

\begin{figure*}[!ht]
\centering
\newlength{\synthfigwidth}
\setlength{\synthfigwidth}{0.33\linewidth}
\begin{subfigure}{\textwidth}
\includegraphics[width=\synthfigwidth]{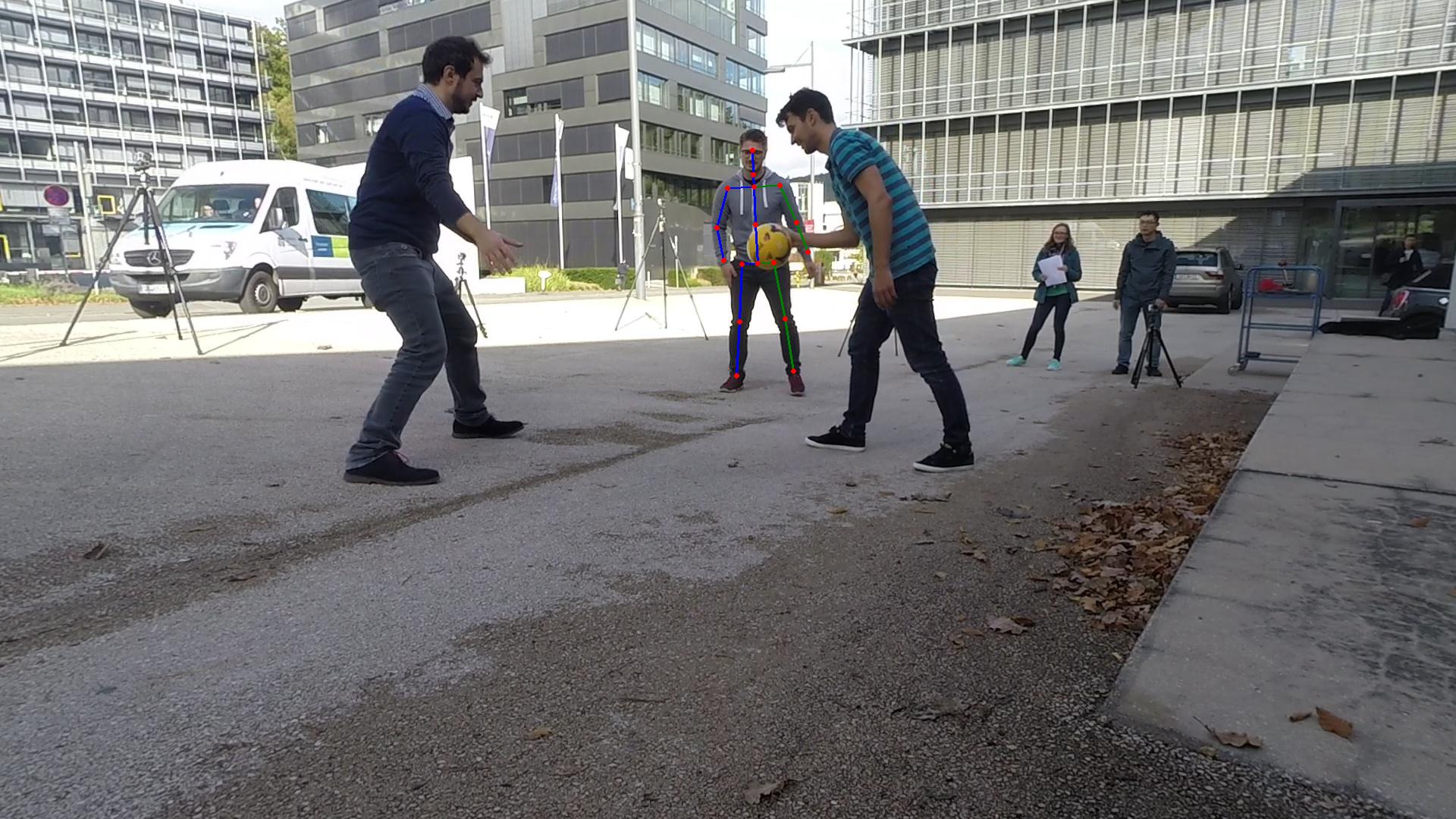}
\includegraphics[width=\synthfigwidth]{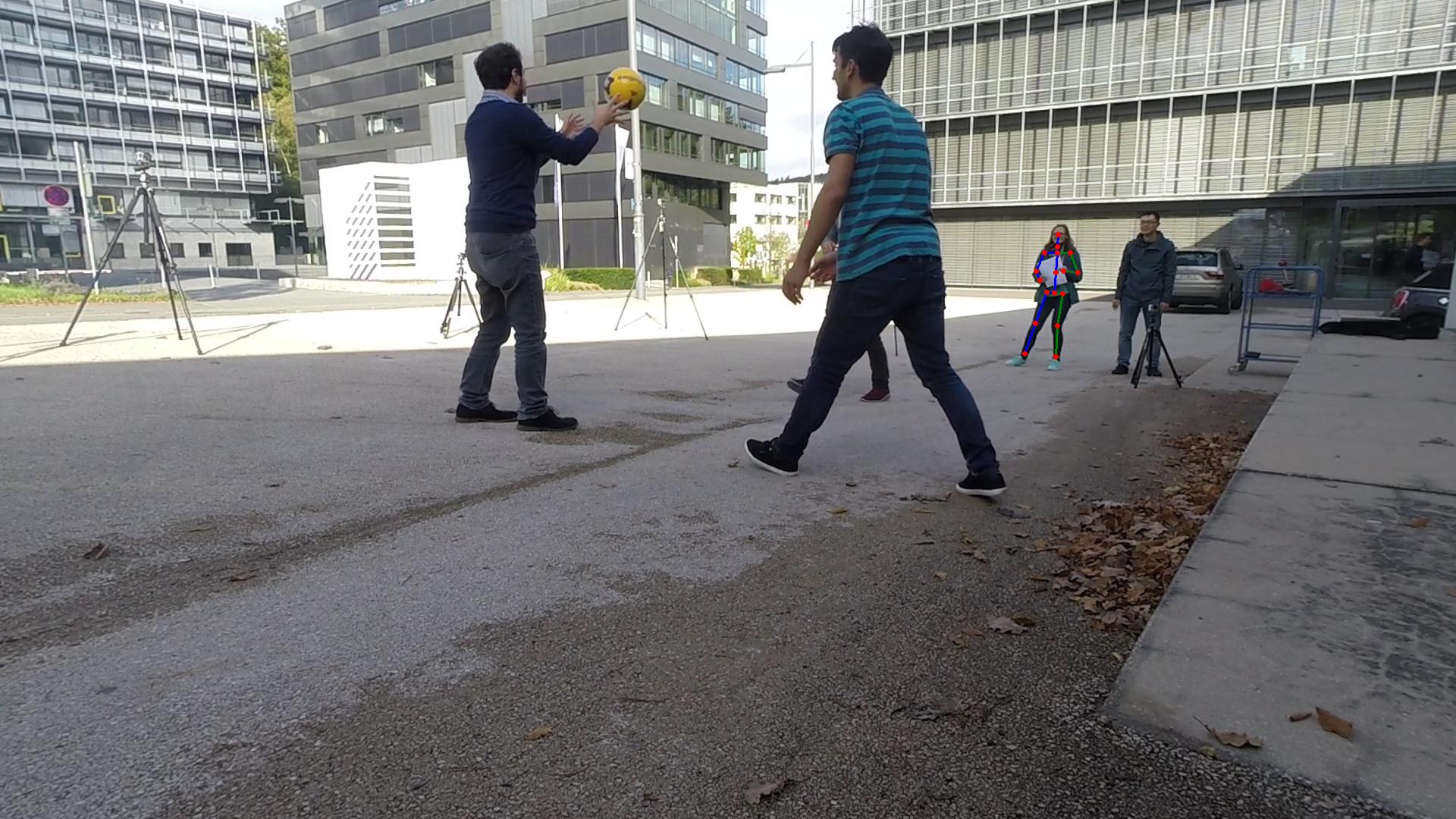}
\includegraphics[width=\synthfigwidth]{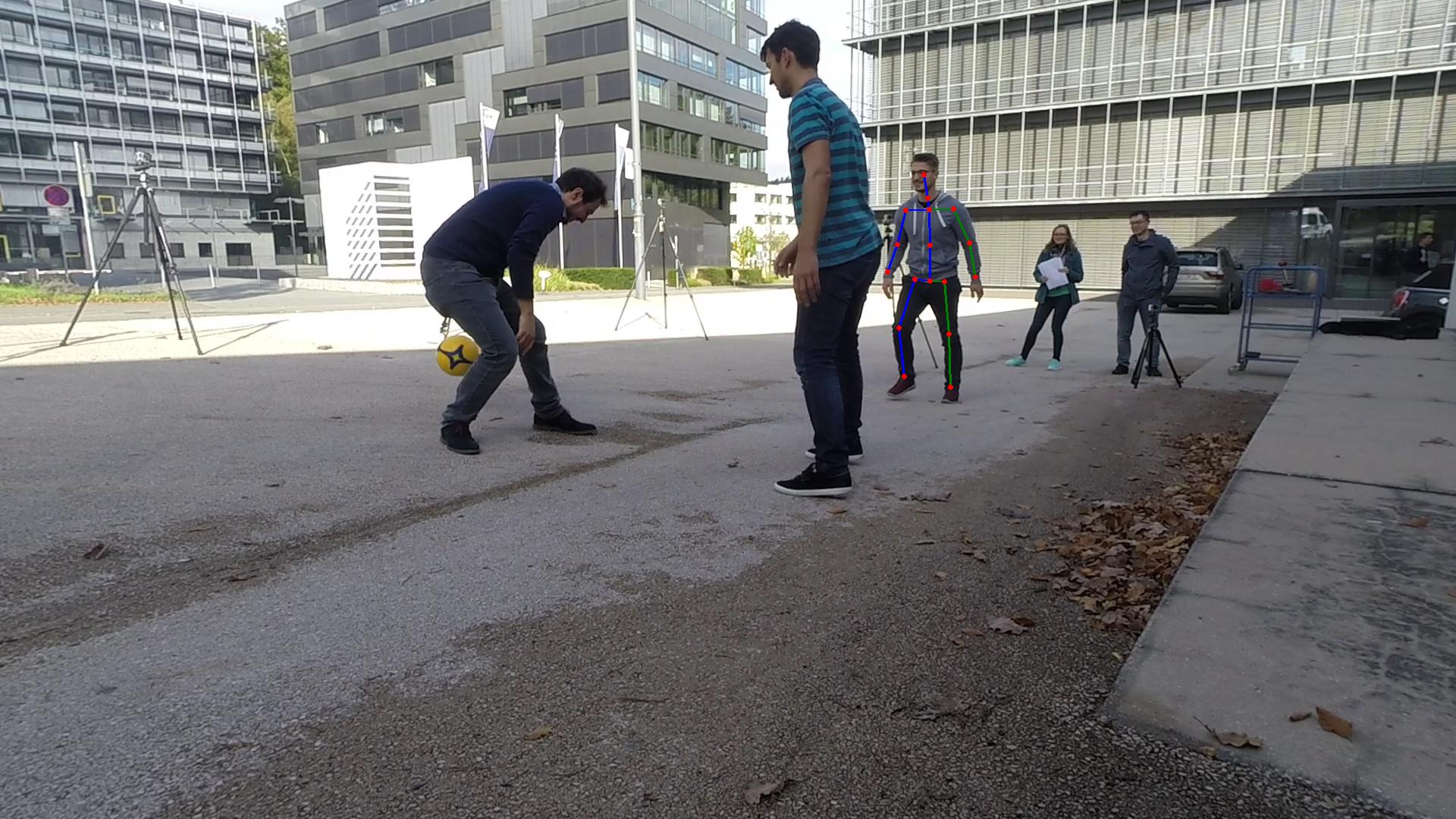}
\caption{\label{fig:xxxxx} PoseBERT input (LCR-Net++). \vspace{0.2cm}}
\end{subfigure}
\begin{subfigure}{\textwidth}
\setlength{\synthfigwidth}{0.33\linewidth}
\includegraphics[width=\synthfigwidth]{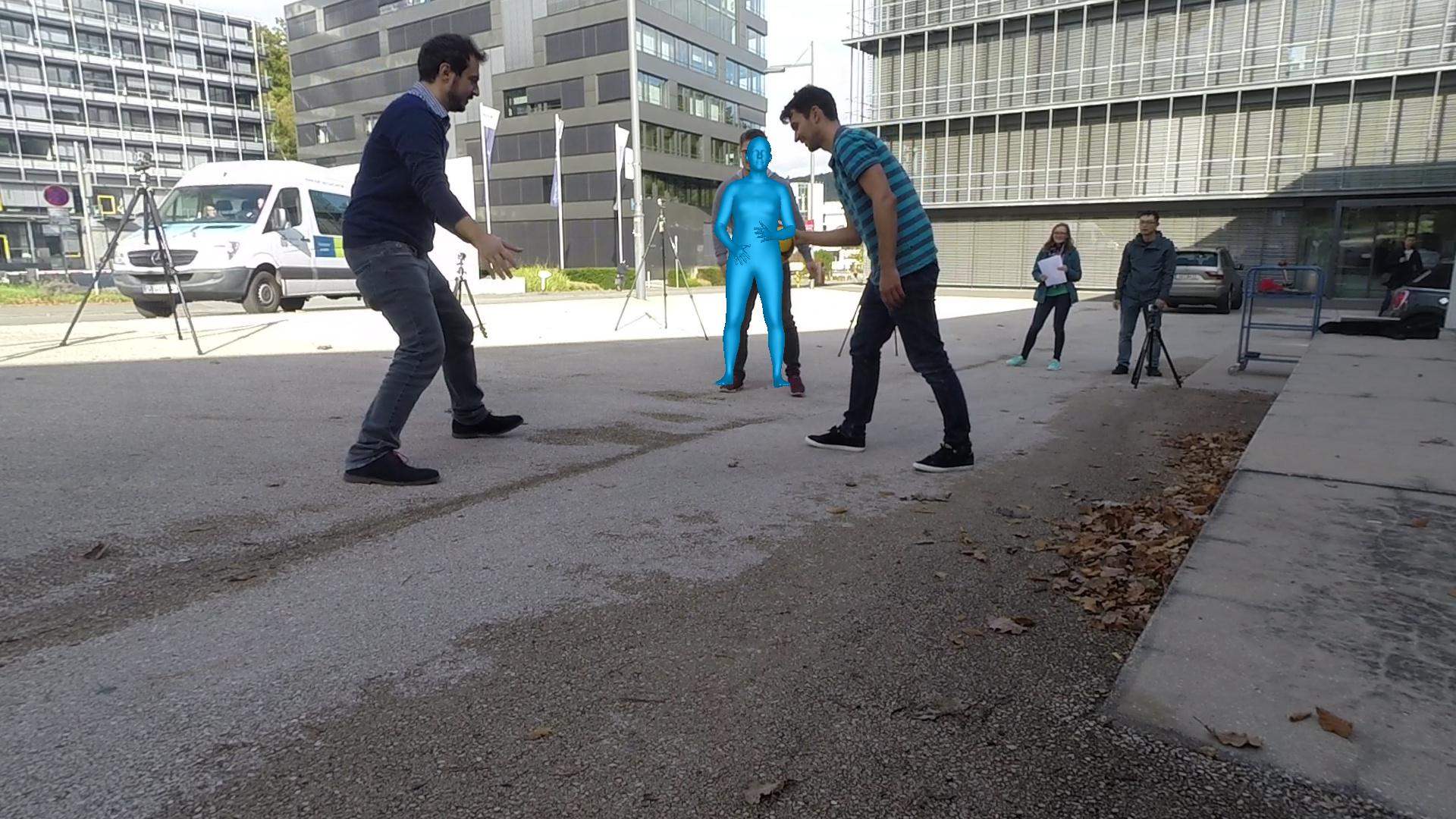}
\includegraphics[width=\synthfigwidth]{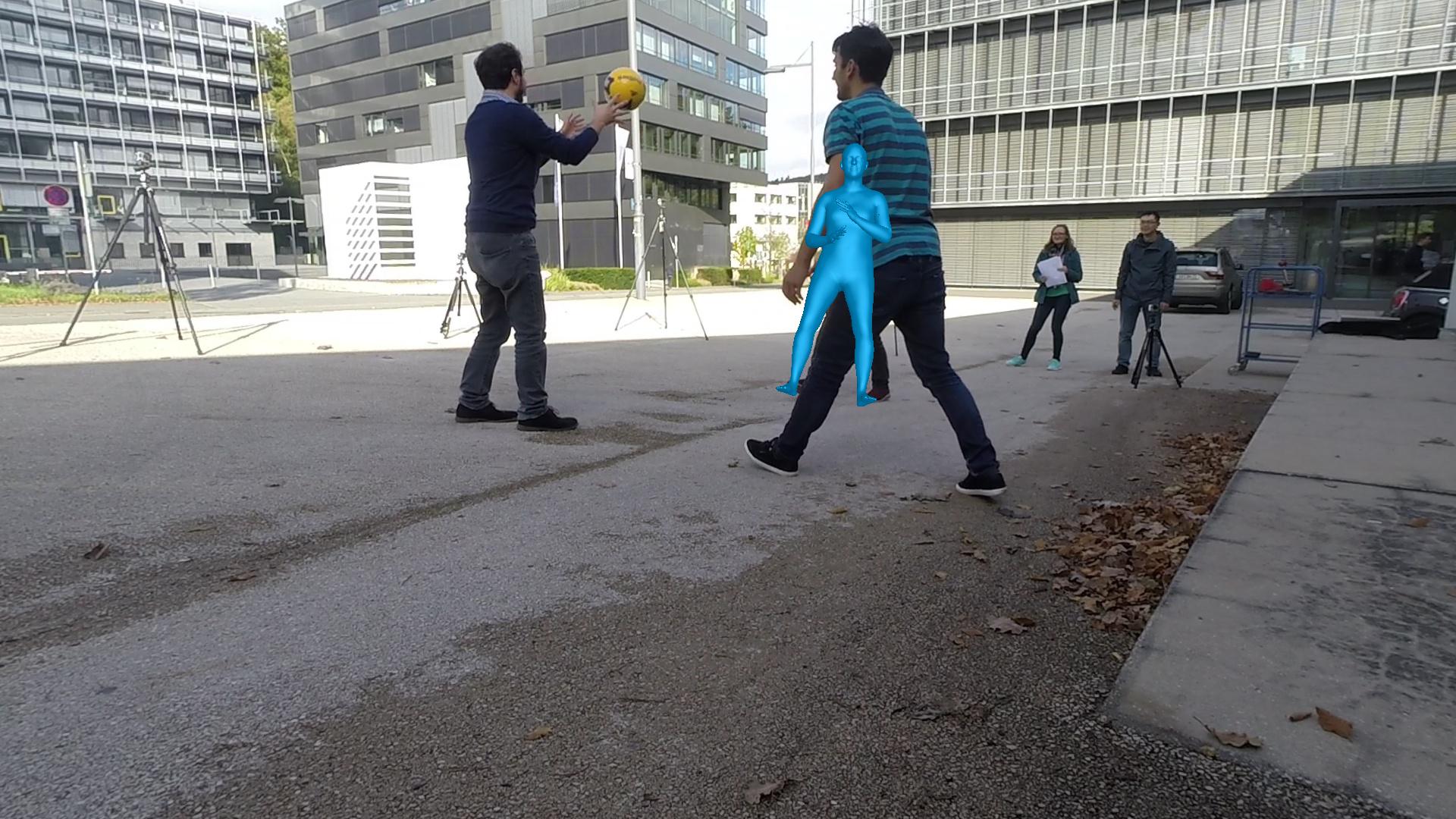} 
\includegraphics[width=\synthfigwidth]{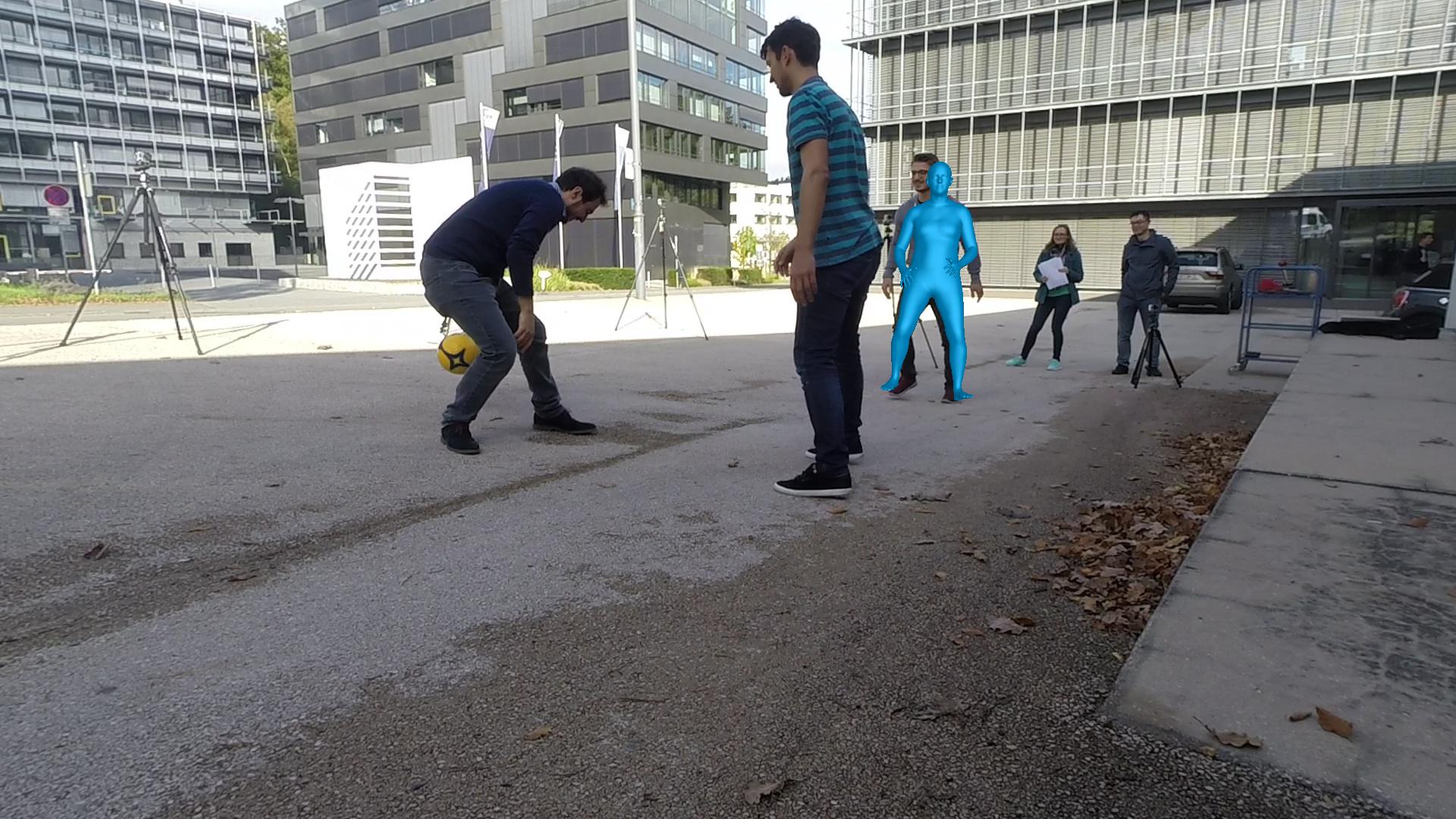}
\caption{\label{fig:xxxx} PoseBERT output.}
\end{subfigure}
\caption{\label{fig:posebert_mupots} \textbf{Qualitative results on the MuPots dataset without using 2D bounding boxes as input.}
Corresponding frames from a sequence depicting the input and output of \posebert for the human body case.
The person of interest is occluded for several consecutive frames.
The image-based model is not able to detect this person however \posebert still produces a plausible sequence of human meshes (cf middle frame).
}
\end{figure*}

We conduct experiments with \posebert using two parametric 3D models: SMPL \cite{SMPL-X:2019} for modeling human body mesh and MANO \cite{romero2017mano} for modeling hand mesh.
In Section \ref{sub:body}, we use \posebert to regress the 3D body pose using the SMPL model~\cite{smpl} assuming 2D ground-truth person bounding box or not.
We conduct an ablation study on the architecture and training strategy of the transformer.
In Section \ref{sub:hand}, we introduce \posebert to model the 3D hand using the MANO model~\cite{romero2017mano}.
We conduct experiments on several tasks ranging from 3D pose estimation to future pose prediction.
Finally, we showcase the use of \posebert in a real-time application of robotic tele-operation.

\paragraph{Training details.}
We train \posebert from scratch using Adam optimizer \cite{adam} with a learning rate of 1e-3 and default parameters.
We train for 2 million iterations on a single V100 GPU.
It takes 2 days to reach this number of iterations but we do not observe a saturation with a big enough network and strong input perturbations coupled with data augmentation.
The framework is implemented using PyTorch \cite{pytorch}.

\subsection{\posebert for human body mesh modeling}
\label{sub:body}
After presenting datasets and metrics in Section~\ref{sub:data}, we study the impact of \posebert on top of existing image-based models in Section~\ref{sub:posebert_smpl_sota}, perform extensive evaluations in Sections~\ref{sub:posebert_smpl_training} and \ref{sub:posebert_smpl_training_bis}, and ablations in Section~\ref{sub:posebert_smpl_hyperparams}.
Finally, we present an evaluation without using ground-truth 2D person bounding box in Section~\ref{sub:beyond_gt_2d_bbox}.

\subsubsection{Datasets, training and metrics}
\label{sub:data}

\noindent \textbf{MoCap data.} We train \posebert solely on MoCap data. We use AMASS~\cite{amass}, which is a collection of numerous Motion Capture datasets in a unified SMPL format, representing more than 45 hours of recording.
We use the training set for training \posebert which contains 11'000 sequences with sequence lengths ranging from few frames to 30 seconds-long sequence.
We downsample the framerate to 30 fps for the entire corpus.



\noindent \textbf{Test datasets and metrics.}
For evaluation, we use the 3DPW~\cite{3dpw} test set, the MPI-INF-3DHP~\cite{mpiinf} test set, the MuPoTS-3D dataset~\cite{mupots} and the AIST dataset~\cite{aist} that contains more challenging poses from people dancing. 
We report the mean per-joint error (MPJPE) before and after procrustes alignment (PA-MPJPE) in millimeters (mm). For 3DPW, following the related work, we also report the mean per-vertex position error (MPVPE). To measure the jittering of the estimations on the video datasets (3DPW, MPI-INF-3DHP, MuPoTS-3D, AIST), we follow~\cite{hmmr} and report the acceleration error, measured as the average difference between ground-truth and predicted acceleration of the joints.
\fbt{
We also report the percentage of correct keypoints in 3D (PCK3D) and its procrustes aligned variant (PA-PCK3D).
A 3D keypoint is said to be correct if it lies within 15cm from the ground-truth keypoint location.
}

\subsubsection{Impact on image-based models}
\label{sub:posebert_smpl_sota}

One major key benefit of \posebert is that it can be plugged on top of any image-based model to transform it into a video-based model since it takes \textit{only} SMPL sequences (for this case) as input compared to other methods such as VIBE~\cite{vibe} or TCMR~\cite{tcmr} which require image-based features as input.

In Table~\ref{tab:posebert_impact}, we report the PA-MPJPE on the 4 video datasets. We observe that when plugging \posebert on top of SPIN~\cite{spin}, it leads to a consistent improvement of 2.3mm on 3DPW, 3.7mm on MPI-INF-3DHP, 2.1mm on MuPoTS-3D and 1.6mm on AIST.
Interestingly, these results are better than the one obtained when using the state-of-the-art video-based method VIBE~\cite{vibe} on MPI-INF-3DHP, MuPoTS-3D and AIST. When using \sspin \cite{leveraging_mocap} as image-based model, we observe a similar consistent improvement on all datasets.

Actually, one can even plug \posebert on top of a model that already leverages videos, such as VIBE, and we observe a similar consistent gain, which suggests that \posebert is complementary to the way temporal consistency of features is exploited in VIBE.



\begin{table}
\centering
\resizebox{\linewidth}{!}{
\begin{tabular}{lllll}
\toprule
 & \multicolumn{1}{c}{3DPW} & \multicolumn{1}{c}{MPI-INF-3DHP} & \multicolumn{1}{c}{MuPoTS-3D} & \multicolumn{1}{c}{AIST} \\
\midrule 
SPIN~\cite{spin} & 59.6	& 68.0 & 83.0 & 76.2 \\
~~~~+ \posebert & \bf{57.3} \errorgain{2.3} & \bf{64.3} \errorgain{3.7} & \bf{80.9} \errorgain{2.1} & \bf{74.6} \errorgain{1.6} \\
\midrule
VIBE~\cite{vibe} & 56.5 & 65.4 & 83.4 & 76.0 \\
~~~~+ \posebert & \bf{54.9} \errorgain{1.6} & \bf{64.4} \errorgain{1.0} & \bf{81.0} \errorgain{2.4} & \bf{74.5} \errorgain{1.5} \\ 
\midrule
\sspin~\cite{leveraging_mocap} & 55.6 & 66.7 & 81.0 & 75.7 \\ 
~~~~+ \posebert & \bf{52.9} \errorgain{2.7} & \bf{63.8} \errorgain{2.9} & \bf{79.9}   \errorgain{1.1} & \bf{74.1} \errorgain{1.6}\\
\bottomrule
\end{tabular}
}


\caption{\textbf{Adding \posebert on top of various methods.} We report the PA-MPJPE metric (lower is better) on four video datasets. The gains in mm are shown in parenthesis.}
\label{tab:posebert_impact}
\end{table}

\subsubsection{Impact of masking and noise perturbation for training}
\label{sub:posebert_smpl_training_bis}

We then study the impact of various training strategies on MoCap datasets in Table~\ref{tab:posebert_pretraining}.
First, we study the impact of partially masking the input sequences, and observe that masking 12.5\%, \ie, 2 frames out of 16, lead to smoother prediction (lower error acceleration) while the PA-MPJPE remains low.

\begin{figure}
    \centering
    \includegraphics[width=\linewidth]{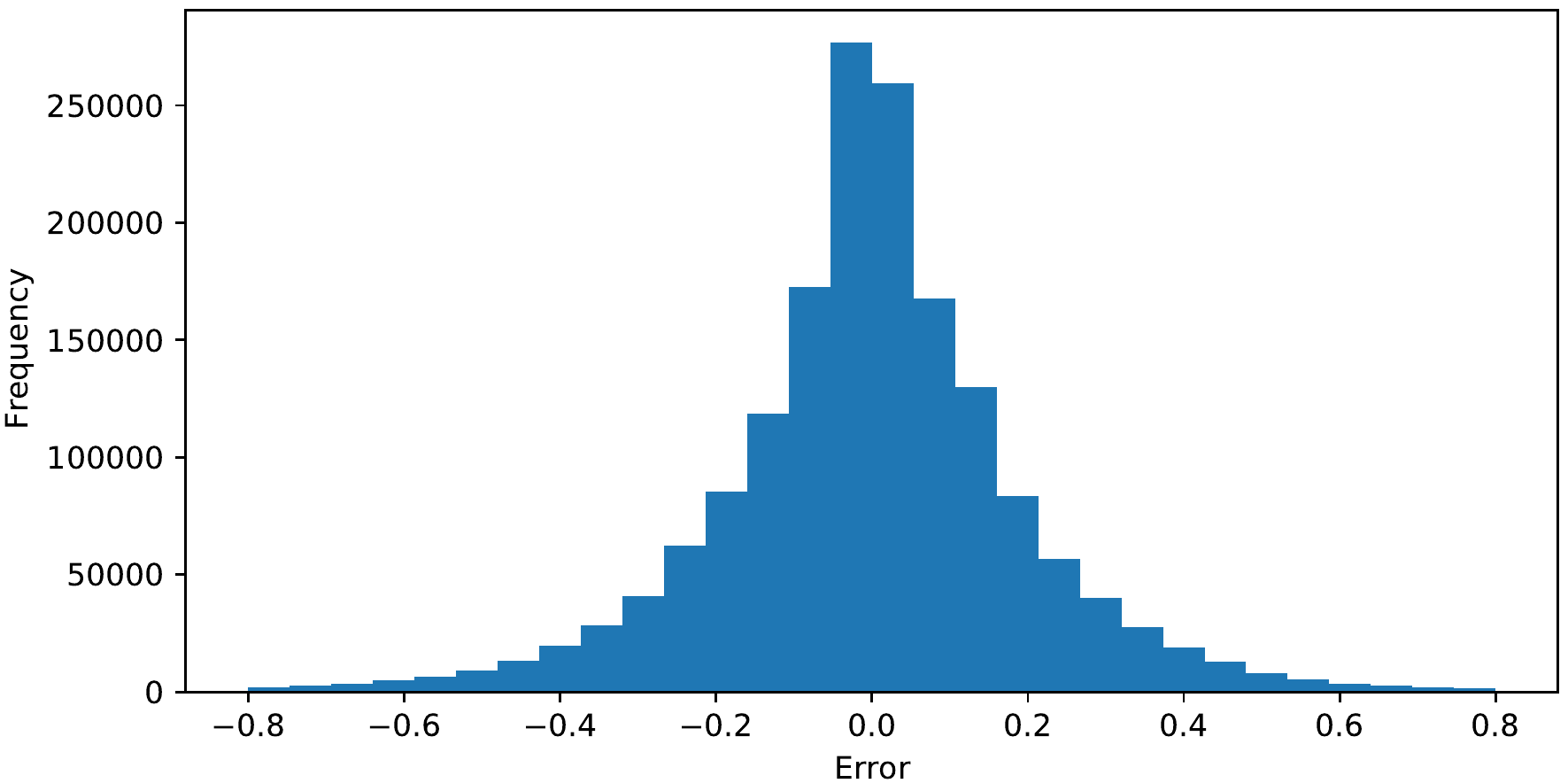}  
    \caption{\textbf{Histogram of SPIN axis-angle errors.} On 3DPW train set, in radians.}
    \label{fig:error_gaussian}
\end{figure}

We also try adding Gaussian noise, with a standard deviation of 0.05 on top of the axis-angle representation, and obtain a small additional boost of performance and smoother predictions.
Increasing the standard deviation did not bring any benefit.
The histogram of the SPIN axis-angle errors in radians shown in Figure \ref{fig:error_gaussian} was  a motivation for adding Gaussian noise.

\begin{table*}
\centering
\resizebox{\textwidth}{!}{
\begin{tabular}{c|l|lllll|llll}
\toprule
 \multirow{2}{*}{2D Detection} & \multirow{2}{*}{Pose Regression} & \multicolumn{5}{c|}{3DPW} & \multicolumn{4}{c}{MuPoTS-3D} \\
  &  & PA-MPJPE $\downarrow$ & PA-PCK3D $\uparrow$ & MPJPE $\downarrow$ & PCK3D $\uparrow$ & MPVPE $\downarrow$ &  PA-MPJPE $\downarrow$ & PA-PCK3D $\uparrow$ & MPJPE $\downarrow$ & PCK3D $\uparrow$ \\
\midrule
\multirow{3}{*}{Ground-truth} & HybrIK~\cite{Li_2021_CVPR} & 47.0&	97.0&	75.2&	91.3&	89.7&	71.6&	94.1&	123.8&	73.5 \\
& ~~~~+ bi-GRU       & 48.0 &	97.2&	83.9&	88.2&	99.1	&71.6&	94.1&	123.6&	73.5\\
& ~~~~+  \bf{\posebert} & \bf{46.2} \errorgain{0.8} &	\bf{97.6} \accgain{0.6} &	\bf{74.6} \errorgain{0.6} &	\bf{91.7} \accgain{0.4} &	\bf{88.9} \errorgain{0.8} &	\bf{71.0} \accgain{0.6} &	\bf{94.6} \errorgain{0.5}&	\bf{122.4} \errorgain{1.4} &	\bf{73.5} \accgain{0.0}\\
\midrule 
\multirow{7}{*}{Faster-RCNN} & HybrIK~\cite{Li_2021_CVPR} & 52.8	&95.2&	98.1&	87.2&	114.9&	76.9&	92.2&	162.1&	71.3 \\
& ~~~~+ bi-GRU       & 50.9&	96.4&	94.0&	85.4&	110.3&	73.9&	93.5&	158.0	&67.1 \\
& ~~~~+ \bf{\posebert} & \bf{47.9} \errorgain{1.8}	&\bf{97.2} \accgain{2.0} &	\bf{76.6} \errorgain{21.5} &	\bf{91.3} \accgain{4.1} &	\bf{91.3} \errorgain{23.6}&	\bf{71.4} \errorgain{5.5}&	\bf{94.5} \accgain{2.3}&	\bf{121.5} \errorgain{40.5}&	\bf{73.5} \accgain{2.2} \\
 & ~~~~+ \bf{\posebert} $m=0$ & 48.1	&97.2&	76.9&	90.7&	91.6&	71.6&	94.5&	121.7&	73.3 \\
& ~~~~+ \bf{\posebert} $m=5$ & 48.3	&97.2&	77.1&	90.6&	92.0&	71.6&	94.5&	121.7	&73.3 \\
& ~~~~+ \bf{\posebert} $m=10$ & 50.3&	96.7&	79.7&	89.5&	95.1&	72.8&	94.2&	123.6	&72.5 \\
& ~~~~+ \bf{\posebert} $m=20$ & 58.0&	94.8&	88.1&	86.0&	105.1&	75.6&	93.1&	126.1&	71.3\\
\bottomrule
\end{tabular}
}
\caption{\fbt{\textbf{Impact of \posebert on top of HybrIK \cite{Li_2021_CVPR}.} 
\rbt{We report results obtained using either ground-truth 2D detections or results of an off-the-shelf algorithm (Faster-RCNN).}
\rbt{We also study the context modeling capabilities of \posebert by masking inputs corresponding to the timestep to predict and the $m$ previous and following timesteps (for $m=0$, only the input of the timestep to predict is masked).}}
\label{tab:posebert_hybrik}}
\end{table*}

\begin{table}
\centering
\resizebox{\linewidth}{!}{
\begin{tabular}{lcc|cc|cc}
\toprule
 & \multicolumn{2}{c|}{3DPW} & \multicolumn{2}{c|}{MPI-INF-3DHP} & \multicolumn{2}{c}{MuPoTS-3D} \\
 Masking \% & PA-MPJPE $\downarrow$ & Accel $\downarrow$ & PA-MPJPE $\downarrow$ & Accel $\downarrow$ & PA-MPJPE $\downarrow$ & Accel $\downarrow$ \\
\midrule
\sspin~\cite{leveraging_mocap} & 55.6 & 32.5 &	66.7 & 29.5 &  81.0 & 23.5 \\
\midrule 
 0\% &    53.3 & 9.6 & \bf{62.3} & 9.8 & 80.0 & 13.8 \\
12.5\% & 53.2 & \bf{7.8}  & 63.8 & \bf{8.7} & 80.3 & \bf{12.8} \\
25\% &   53.3 & 8.3 & 64.2 & 9.0 & 80.3 & 13.3 \\
37.5\% & 53.9 & 9.0 & 65.0 & 9.2 & 80.8 & 14.0 \\
\midrule
12.5\% + Noise & \bf{52.9} & 8.3 & 63.3 & \bf{8.7} & \bf{79.9} & 13.7 \\
\bottomrule
\end{tabular}
}
\caption{\textbf{Ablation on the \posebert pretraining strategy.} We study the impact of masking the input sequences and adding Gaussian noise.}
\label{tab:posebert_pretraining}
\end{table}

\subsubsection{Additional ablation study of the training strategy}
\label{sub:posebert_smpl_training}
In addition to masking and adding Gaussian noise on the input, we have also investigated other training strategies as reported in Table~\ref{tab:posebert_ablation_2}.

First we compare against the common practice of having the iterative regressor \cite{thmmr} on top of the temporal module.
\posebert shows a gain ranging from 1.4mm to 0.4mm compared to the baseline described above.

We then increase the temporal window of the input sequence by reducing the frames per second while keeping the sequence length fixed.

We observe that increasing the time span does not bring significant improvement and even leads to decreased performances.
We also study the impact of incorporating random poses or joints compared to random Gaussian noise.
We note that both noise types bring a small improvement compared to Gaussian noise but for simplicity we do not include them during the training scheme of our best model.

\begin{table}[h]
\centering
\resizebox{\linewidth}{!}{
\begin{tabular}{lccc}
\toprule
 & 3DPW & MPI-INF-3DHP & MuPoTS-3D \\
\midrule
\sspin~\cite{leveraging_mocap} & 55.6 & 66.7	& 81.0 \\ 
\midrule
\sspin + Transformer + Regressor & 54.5 & 65.2 & 80.7 \\
\midrule
\sspin + \bf{\posebert} & \bf{53.2} & 63.8 & \bf{80.3} \\
~~~~fps=7.5 & 54.0 & 64.5 & 80.5 \\
~~~~fps=15 & 53.4 & \bf{63.4} & 80.4 \\
~~~~fps=3.75 & 55.0 & 66.0 & 81.0 \\
\midrule
~~~~Random poses ~~5\% & 53.2 & \bf{63.5} & \bf{80.0} \\
~~~~~~~~~~~~~~~~~~~~~~~~~~~~ 10\%  & \bf{53.2} & 63.7 & \bf{80.0} \\
~~~~~~~~~~~~~~~~~~~~~~~~~~~~ 20\% & \bf{53.2} & 64.3 &  80.2 \\
~~~~~~~~~~~~~~~~~~~~~~~~~~~~ 40\%  & 53.6 & 64.5 &  80.4\\
\midrule 
~~~~Random joints ~~5\% & \bf{53.2} & \bf{62.7} & \bf{80.1} \\
~~~~~~~~~~~~~~~~~~~~~~~~~~~~ 10\%  & 53.4 & 63.6 & 80.2 \\
~~~~~~~~~~~~~~~~~~~~~~~~~~~~ 20\% & 53.3 & 64.9 & 81.0 \\
~~~~~~~~~~~~~~~~~~~~~~~~~~~~ 40\%  & 55.8 & 70.0 & 83.0 \\
\bottomrule
\end{tabular}
} 
\caption{\textbf{Additional ablation on the \posebert hyperparameters.} We first study the impact of having the regressor incorporated into the transformer. We also study the impact of the frame per second (fps, 30 by default) and the percentage of random poses/joints (0 by default) with the PA-MPJPE metric on 3DPW, MPI-INF-3DHP and MuPoTS-3D when using \posebert on top of \sspin~\cite{leveraging_mocap}, with masking 12.5\% of the input sequences (2 frames with T=16 frames) and using a model of size $D=512$ and $L=4$. 
}
\label{tab:posebert_ablation_2}
\end{table}

\begin{figure*}[!ht]
\centering
\setlength{\synthfigwidth}{0.33\linewidth}
\begin{subfigure}{\textwidth}
\includegraphics[width=\synthfigwidth]{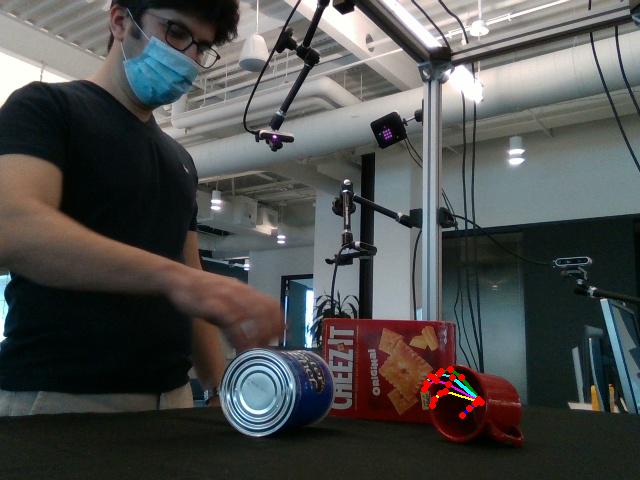}
\includegraphics[width=\synthfigwidth]{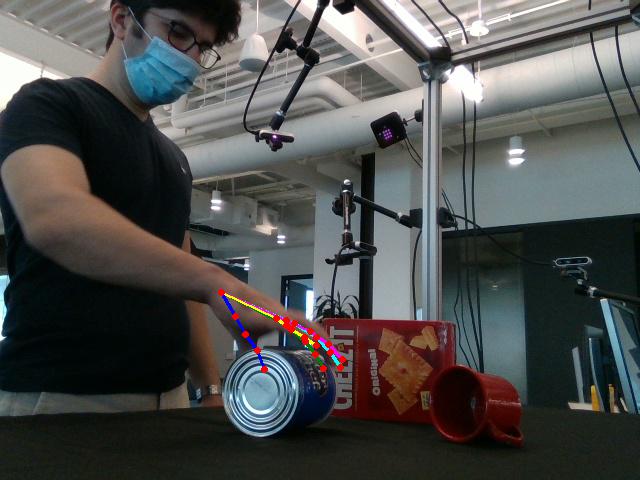}
\includegraphics[width=\synthfigwidth]{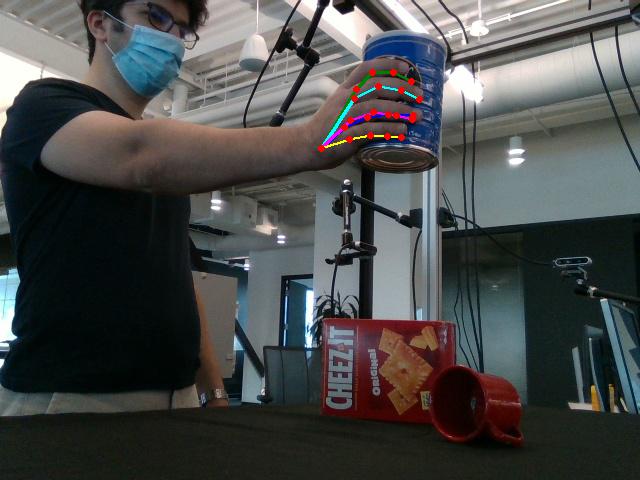}
\caption{\label{fig:xxxxx} PoseBERT input (LCR-Net hand expert). \vspace{0.2cm}}
\end{subfigure}
\begin{subfigure}{\textwidth}
\setlength{\synthfigwidth}{0.33\linewidth}
\includegraphics[width=\synthfigwidth]{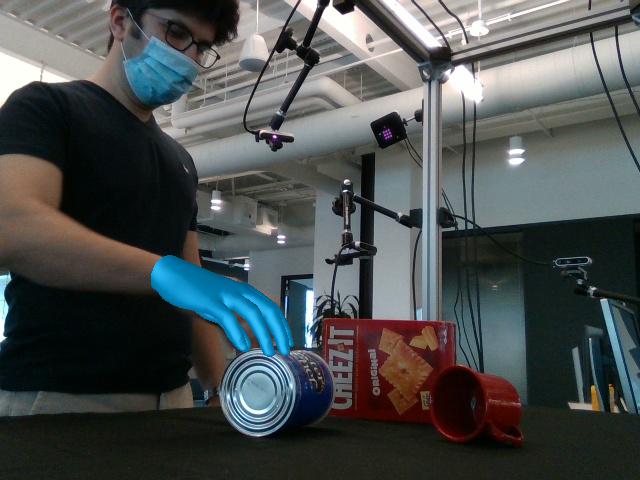}
\includegraphics[width=\synthfigwidth]{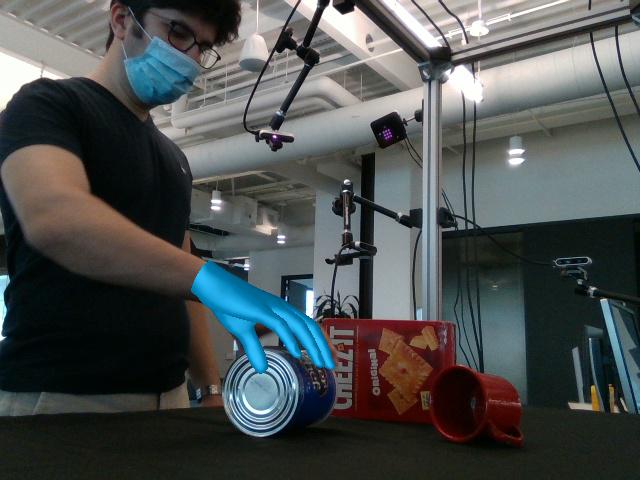} 
\includegraphics[width=\synthfigwidth]{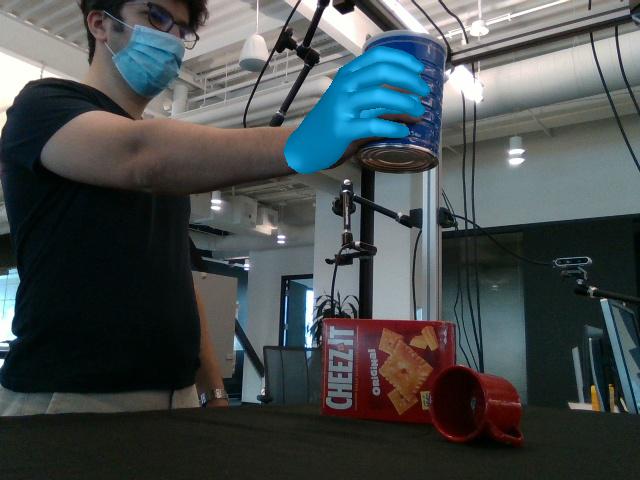}
\caption{\label{fig:xxxx} PoseBERT output.}
\end{subfigure}
\caption{\label{fig:posebert_DexYCB_2} \textbf{Qualitative results on the DexYCB dataset.}
Corresponding frames from a sequence depicting the input and output of \posebert for the hand expert case.
Note that for the LCR-Net estimation of the first frame of the video is totally wrong (motion blur), however \posebert is able to recover a plausible pose using the temporal information.
}
\end{figure*}

\subsubsection{Impact of hyperparameters}
\label{sub:posebert_smpl_hyperparams}

As a first step we study the impact of some architecture design choices and report the PA-MPJPE on three datasets in Table~\ref{tab:posebert_ablation}.
As a default training strategy we mask 12.5\% of the input poses for this ablation.
First we note that \posebert leads to a consistent gain of 1 to 3mm on all datasets.
Removing the positional encoding leads to a suboptimal performance indicating that incorporating temporal information within the network is a key design choice.
Sharing the regressor allows to reduce the number of learnable parameters and leads to better predictions.
More importantly we notice that we can iterate over the regressor only a single time after each layer given that doing more iterations does not improve and even slightly decreases the performance.
In terms of model size, the benefit of \posebert seems to be reached with a depth of $L=4$ and an embedding dimension of $D_t=512$.
We choose these hyperparameters since increasing the model complexity leads to minimal improvements.
For the temporal length of the training sequence, we set $T=16$, as longer sequences do not lead to further improvements.
With such settings, using \posebert introduces a limited overhead of 7.3M parameters and 0.13GFLOP per sequence (0.53GFLOP for $T=64$),
compared to the 27.0M parameters and 4.2GFLOP per image required by the SPIN image-based model.

\subsubsection{Beyond the use of ground-truth 2D person bounding boxes. }
\label{sub:beyond_gt_2d_bbox}
For the experiments presented before we follow state-of-the-art setup and use the ground-truth 2D bounding box for cropping around the person.
Following this scenario, there is no missing detections and noisy image-based estimations happen only when the person is strongly occluded.
To consider a more realistic scenario, we perform experiments without using the 2D ground-truth bounding boxes.
We take \textit{only} the entire RGB video as input and run LCR-Net++ \cite{lcrnet++} to detect people and estimate their associated 3D pose.
LCR-Net++ is a multi-person 2D-3D human pose detector which is robust to in-the-wild scenarios and runs in real-time.
We do a per-frame detections assignment using the Hungarian matching of the predicted 2D poses.
On top of that, we build the sequences of poses using the ground-truth person identity.
LCR-Net++ is able to assign a pose to 94\% of the persons (with ground-truth annotations) in MuPoTS.

We observed that for this setup, \posebert need to be trained on longer sequences since the image-based model produces noisier and more incomplete sequences.
Thus, taking into account a larger temporal window is mandatory to update the initial pose with more contextual information.
We set the sequence length to 128 frames.

We report results in Table~\ref{tab:posebert_mupots}.
We observe that \posebert brings a significant boost on all metrics compared to LCR-Net++ only (more than 20 mm MPJPE and PA-MPJPE).
We also report a temporal filtering baseline based on the Savitzky-Golay filter~\cite{Savitzky-Golay} as in \cite{MetaFuse}.
In this case, the missing poses are replaced by the nearest ones and we apply the smoothing method on the sequence.
The Savitzky-Golay filtering marginally improves  the results but is not able to correct implausible initial poses.
Compared to this smoothing baseline, \posebert shows a significant gain of 11.92 mm (resp. 3.97 ) on MPJPE (resp. PA-MPJPE).

Beyond improving the overall 3D poses it is also interesting to notice that \posebert outputs smooth human mesh such as demonstrated by the Accel metric.

In Figure~\ref{fig:posebert_mupots}, we show a qualitative results of \posebert when adding on top of LCR-Net++ outputs.
In the middle frame, we observe that LCR-Net++ is not able to provide a pose estimate for the person of interest.
It is mainly due to a strong occlusion with the person in front.
This occlusion remains very important for a few frames.
\posebert is able to provide a plausible 3D human mesh given the LCR-Net++ outputs before and after the occlusion occurs.

\fbt{
\subsubsection{\rbt{Impact of detection quality}}
\label{sub:study_bboxes}

\rbt{Most state-of-the-art methods recovering the 3D pose of a person from an image proceed in two steps. First the person is detected in the image (using an off-the-shelf detector or some ground-truth annotations) and the image is cropped to a 2D bounding box around the person.
Then pose parameters are predicted from this image crop.
In this section, we study the impact of person detection on the overall pose recovery performances, and how PoseBERT can provide improvements by exploiting temporal information.
We use HybrIK~\cite{Li_2021_CVPR} to predict a 3D human pose from a crop, and we compare results obtained using 2D bounding boxes obtained either using Faster-RCNN~\cite{girshick2015fast} or ground-truth annotations.
}
We report results in Table \ref{tab:posebert_hybrik}.

\rbt{When ground-truth bounding boxes are used, an initial image-based prediction is produced for most timesteps of the sequences and the temporal module ``only'' needs to correct these predictions.}
In this 
\rbt{ideal} setting, \posebert brings improvement for all metrics on 3DPW and MuPots-3D, compared to the image-based baseline.
While the 
\rbt{numerical improvement is small,}
it is interesting to note that a recurrent network, 
\rbt{composed of a} stack of bidirectional GRU (biGRU), with a similar model capacity (same number of parameters and input embedding dimension) and same training strategy, performs worst than the image-based model.
\rbt{This suggests that recurrent networks, whose hidden states are updated in a recursive manner, may not be able to generalize at test time to a masking ratio different from the one used during training (number of masked inputs close to 0 in this setting \emph{vs.} 12.5\% during training).}

We then 
\rbt{consider a more realistic scenario using an off-the-shelf algorithm to detect and extract human bounding boxes.}
We use Faster-RCNN~\cite{girshick2015fast} with a score threshold at 0.9,
\rbt{and we consider a person to be well detected if its bounding box has an \emph{intersection-over-union} with the ground-truth of at least 0.5.}
We observe that respectively only 72.3\% and 75\% of the persons are \rbt{well} detected in 3DPW and MuPots-3D, which indicates that deploying a video-based method to fill in missing 3D poses makes a lot of sense for real-world applications.
\rbt{
For evaluation purposes, we replace missing predictions by the temporally closest ones before computing metrics for the image-based method HybrIK.
Yet, we observe that its performance decreases a lot compared to the first scenario.
}
While the bi-GRU baseline brings some improvements compared to the image-based model
\rbt{, these gains are small compared to the ones obtained using \posebert as temporal module.}
In terms of MPJPE, \posebert improves the performance by 21.5 and 40.5 points on 3DPW and MuPots-3D, respectively.
\rbt{Most importantly, performances obtained using \posebert in this setting are close to those obtained assuming ground-truth person bounding boxes.}
It demonstrates the effectiveness and robustness of \posebert 
\rbt{ as a plug-and-play module to deal with misdetections or noisy detections in real world scenarios.}

\subsubsection{Impact of the temporal context for missing timesteps }
\label{sub:study_context}
We also propose a study to better understand how the context is taken into account when \posebert needs to fill in missing 
\rbt{image-based predictions}.
To do so, for each timestep 
\rbt{we mask the input pose corresponding to this timestep as well as the inputs corresponding to the $m$ previous and $m$ following timesteps. We then try to estimate the 3D pose using only the remaining temporal context}
\rbt{(last four rows of Table \ref{tab:posebert_hybrik})}.

First, we mask only the timestep of interest (\ie, $m=0$) and observe that there is a marginal drop in performance. The MPJPE drops from 76.6 to 76.9 on 3DPW and from 71.4 to 71.6 on MuPots-3D.
\rbt{This confirms than \posebert exploits the surrounding timesteps to estimate 3D poses.}

\rbt{If we further increase the masking duration, we observe a bigger drop of performance, but \posebert still outperforms the image-based model even with 10 frames masked around the timestep of interest (\ie, $m=10$).}
\rbt{This suggests that it is able to properly exploit the contextual information, and that filling in the missing detections can still be effective when those are substantial in a sequence.}
}

\begin{figure*}[!th]
\centering
\begin{subfigure}{\textwidth}
\includegraphics[width=\textwidth]{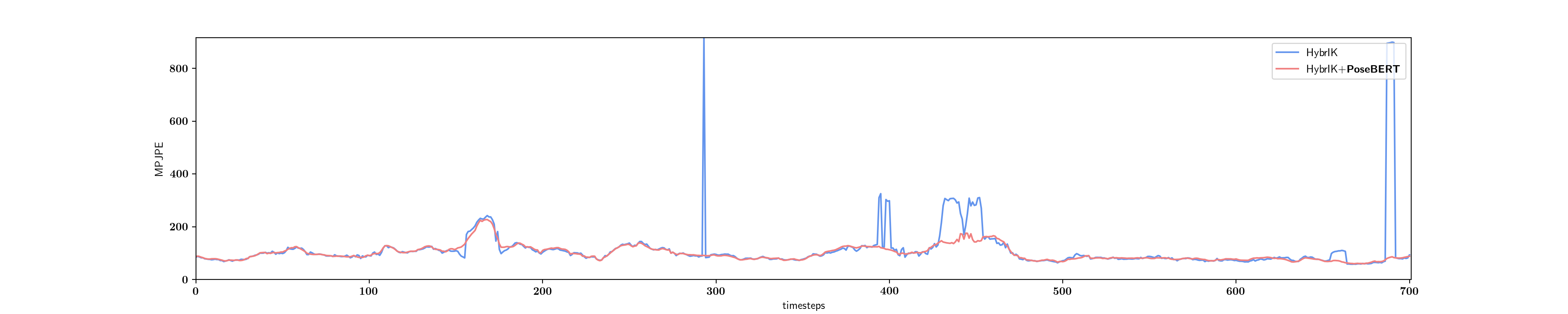}
\caption{\label{fig:xxxx} On MuPots-3D.}
\end{subfigure}
\begin{subfigure}{\textwidth}
\includegraphics[width=\textwidth]{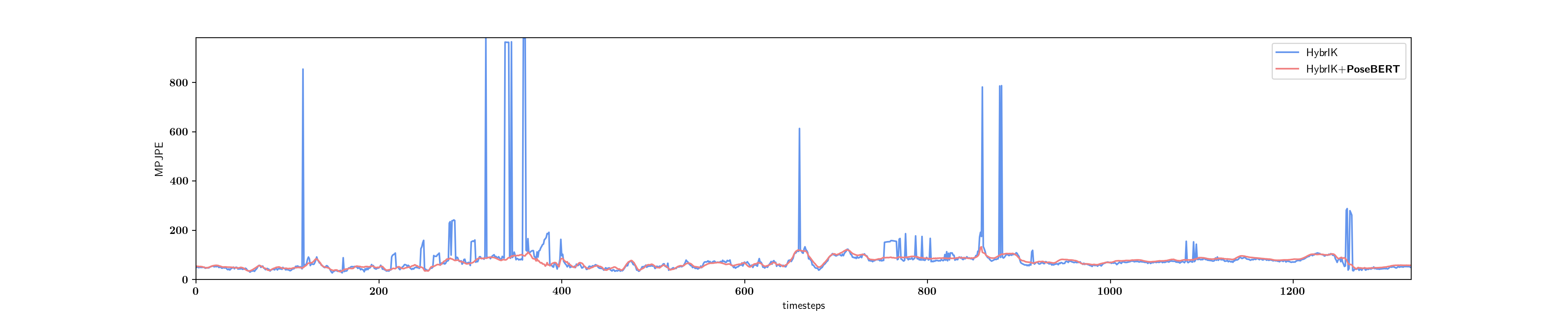}
\caption{\label{fig:xxxx} On 3DPW.}
\end{subfigure}
\caption{\label{fig:posebert_plot_mpjpe}
\fbt{
\textbf{Evolution of the MPJPE on an entire sequence.}
We plot the evolution of the MPJPE across time for HybrIK and HybrIK+\posebert. The detection stage is done using Faster-RCNN.
}
}
\end{figure*}

\fbt{
\subsubsection{\rbt{Finer analysis of \posebert improvements} }
Finally, we study how \posebert is improving the performance of the human mesh recovery pipeline compared to 
\rbt{solely} using the image-based model HybrIK.
\rbt{We evaluate how often \posebert returns a better estimation than HybrIK in term of MPJPE.}
\rbt{When using ground-truth bounding boxes as input, \posebert returns better results than HybrIK only 47.3\% and 46.7\% of the time respectively on 3DPW and MuPots-3D. When using bounding boxes predicted by Faster-RCNN instead,}
the percentages increase to 52.4\% and 50.2\% respectively for 3DPW and MuPots-3D.
To further study how \posebert impacts the final predictions, we plot in Figure \ref{fig:posebert_plot_mpjpe} the 
\rbt{per-frame MPJPE for sequences of 3DPW and MuPots-3D.}
\rbt{We conclude that \posebert is able to recover from large errors made by the image-based method by leveraging the contextual information.}
\rbt{It regularly leads to a small drop of performance however, which happens especially when the performance of the image-based method is already good.}

}



\begin{figure*}[!th]
\centering
\setlength{\synthfigwidth}{0.33\linewidth}
\begin{subfigure}{\textwidth}
\includegraphics[width=\synthfigwidth]{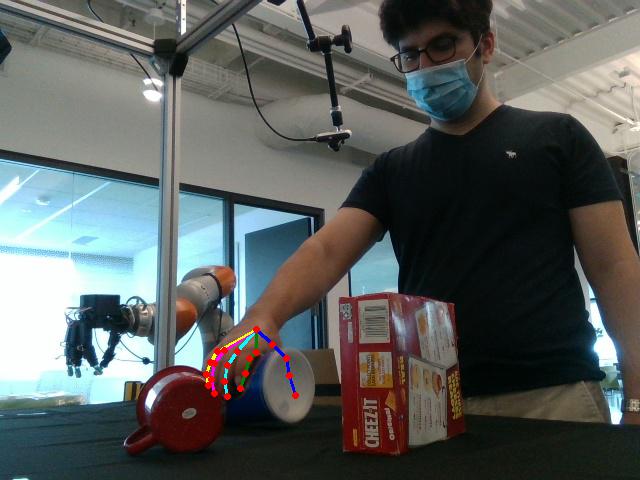}
\includegraphics[width=\synthfigwidth]{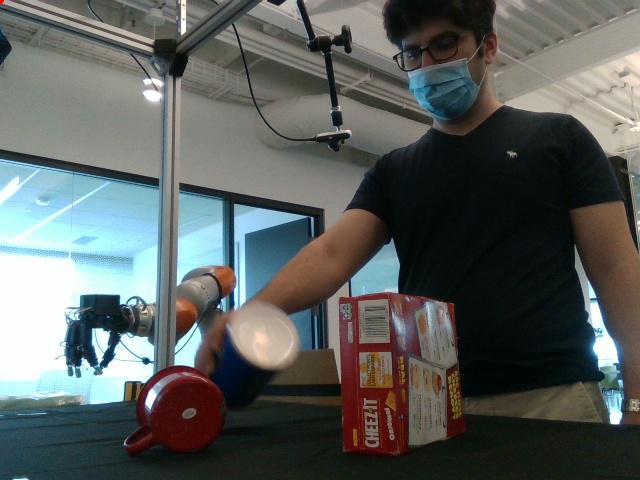}
\includegraphics[width=\synthfigwidth]{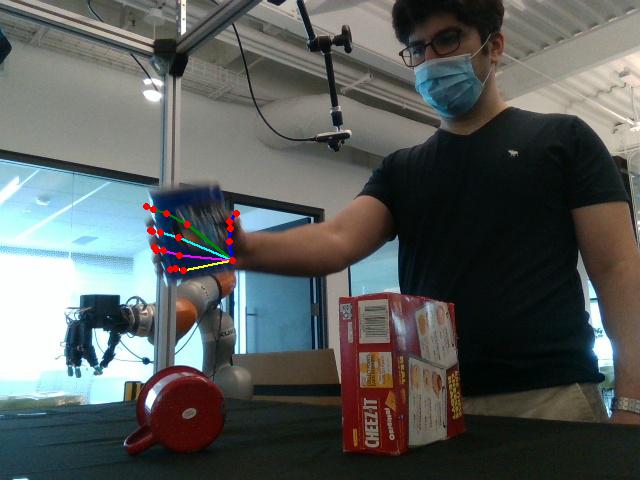}
\caption{\label{fig:xxxxx} PoseBERT input (LCR-Net hand expert).}
\end{subfigure}
\begin{subfigure}{\textwidth}
\setlength{\synthfigwidth}{0.33\linewidth}
\includegraphics[width=\synthfigwidth]{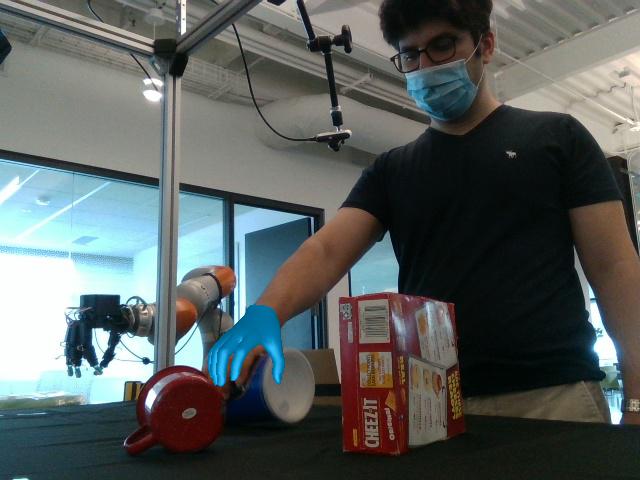}
\includegraphics[width=\synthfigwidth]{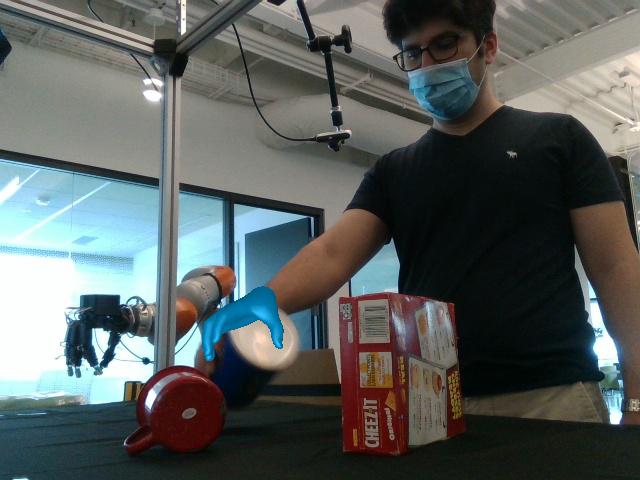} 
\includegraphics[width=\synthfigwidth]{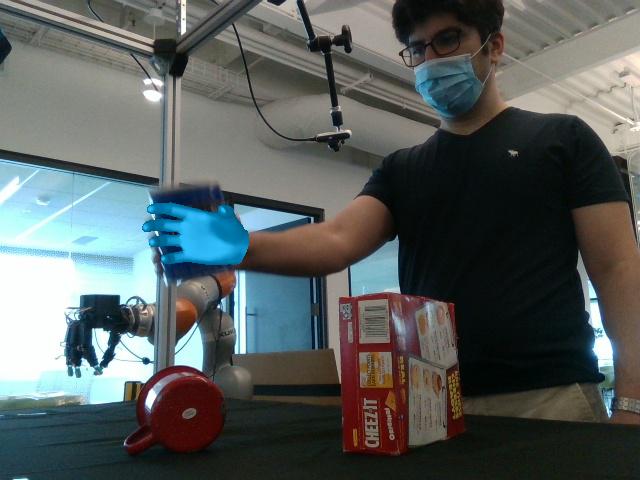}
\caption{\label{fig:xxxx} PoseBERT output.}
\end{subfigure}
\caption{\label{fig:posebert_DexYCB_1} \textbf{Qualitative results on the DexYCB dataset.}
Corresponding frames from a sequence depicting the input and output of \posebert for the hand expert case.
Note that for the middle frame, the LCR-Net prediction is missing due to an heavy occlusion of the hand by the manipulated object.
}
\end{figure*}

\begin{table}
\centering
\resizebox{\linewidth}{!}{
\begin{tabular}{lccc}
\toprule
 & 3DPW & MPI-INF-3DHP & MuPoTS-3D \\
\midrule
\sspin \cite{leveraging_mocap} & 55.6 & 66.7	& 81.0 \\ 
\midrule
\sspin + \bf{\posebert} & \bf{53.2} & \bf{63.8} & \bf{79.9} \\
~~~~w/o pos. encoding & 54.8 & 64.0 & 80.8 \\
~~~~w/o shared regressor & 54.0 & 65.0 & 81.0 \\
~~~~with 2 regressor iterations & 53.3 & 64.0 & 80.5 \\
~~~~with 4 regressor iterations & 53.4 & 64.2 & 80.5 \\
\midrule
~~~~L=1 & 54.5 &	67.0 &	81.2 \\
~~~~L=2 & 53.9 & 	65.5 & 	81.0 \\
~~~~L=4 (default) & \bf{53.2} & 63.8 & 79.9 \\
~~~~L=8 & 53.4 &	\bf{63.3} &	\bf{79.8} \\
\midrule 
~~~~$D_t$=128	& 55.6	& 69.4 &	82.0 \\
~~~~$D_t$=256	& 53.7	& 64.0 &	80.4 \\
~~~~$D_t$=512 (default)	& \bf{53.2} &	63.8	& \bf{79.9} \\
~~~~$D_t$=1024	& 53.4	& \bf{62.9} 	& \bf{79.9} \\
\midrule
~~~~T=8 & 53.9	& 65.2 &	\bf{79.8} \\
~~~~T=16 (default) & \bf{53.2}	& 63.8 &	79.9 \\
~~~~T=32 & 53.4 &	\bf{63.6} & 80.3 \\
~~~~T=64 & 53.3 & 63.7 & 80.4 \\
\bottomrule
\end{tabular}
} 


\caption{\textbf{Ablation on the \posebert hyperparameters.} We study the impact of the positional encoding, sharing the regressor, the number of regressor iterations per layer (1 by default), the depth L of the network (4 by default), the number of channels ($D_t$=512 by default) and the length of the sequences (T=16 by default) with the PA-MPJPE metric on 3DPW, MPI-INF-3DHP and MuPoTS-3D when using \posebert on top of \sspin, with masking 12.5\% of the input sequences (2 frames with T=16 frames).}
\label{tab:posebert_ablation}
\end{table}

\begin{table}
\centering
\resizebox{\linewidth}{!}{
\begin{tabular}{lllll}
\toprule
 Method & MPJPE $\downarrow$ & PA-MPJPE $\downarrow$ & Accel $\downarrow$
 \\
\midrule 
LCR-Net++ ~\cite{lcrnet++} & 153.76  & 105.23 & 37.98 \\
~~~~ \textit{(matched groundtruths only)} & \textit{136.79} & \textit{85.53} & \textit{32.86} \\
~~~~ \textit{(miss detections replaced by nearest detection)} &  \textit{139.36} & \textit{86.42} & \textit{28.25} \\
~~~~~~~~ \textit{(+ Savitzky-Golay filtering)} &  \textit{138.54} & \textit{86.50} & \textit{16.10} \\
~~~~ + \bf{\posebert} & \bf{126.62} \errorgain{27.14} & \bf{82.53} \errorgain{22.70} & \bf{12.78} \errorgain{25.20} \\ 
\bottomrule
\end{tabular}
}
\caption{\textbf{Results on MuPots without using ground-truth 2D bounding boxes as input.} We add \posebert on top of LCR-Net++, a human pose detector. Results are reported in millimeters.}
\label{tab:posebert_mupots}
\end{table}

\subsection{\posebert hand mesh modeling}
\label{sub:hand}
We first present the dataset and metrics in Section \ref{sub:data_hand} and then study the impact of \posebert on top of off-the-shell image-based methods in Section \ref{sub:sota_hand}.
We perform extensive ablations study in Section \ref{sub:ablation_hand} and show results in future hand pose prediction in Section \ref{sub:hand_future}.
Finally we show an application to robotic teleoperation in Section \ref{sub:hand_application}.

\subsubsection{Datasets, training and metrics}
\label{sub:data_hand}

\noindent \textbf{Data.} We use the DexYCB \cite{chao:cvpr2021} dataset for benchmarking \posebert.
This dataset contains 1000 sequences of hand grasping of objects and on average the sequence length is 60 frames.
They are 4 different train/val/test splits. We us the default s0 train/val/test split for our experiments.
For training, we supervise \posebert using the fits of the MANO model to the s0 training set.

\noindent \textbf{Metrics.} We use the same metrics as described in the previous section such as MPJPE, PA-MPJPE and Accel.
We compute these metrics from the 21 keypoints associated with the MANO model.

\subsubsection{Impact on top of image-based models}
\label{sub:sota_hand}

In Table \ref{tab:posebert_impact_hand}, we report state-of-the-art results of two image-based models HR-Net \cite{spurr2020eccv} which was re-implemented by \cite{chao:cvpr2021} and a version of LCR-Net~\cite{lcrnet++} retrained for hand pose detection which was presented in~\cite{hand_challenge2019} and employed as a hand expert in~\cite{dope}. We refer to this second model as LCR-Net - Hand expert.

First, HR-Net \cite{chao:cvpr2021} takes as input the ground-truth hand bounding box and is trained on the DexYCB training set.
Plugging \posebert on top allows to decrease the MPJPE (resp. PA-MPJPE) by 3.29 mm (resp. 2.79 mm). Moreover the decrease in the ``Accel'' metrics from 12.77 to 3.62 demonstrates that \posebert produces much smoother hand pose sequences.

Second, we plug \posebert on top of LCR-Net - Hand expert,  a more realistic image-based model, which is not trained on DexYCB training set. LCR-Net - Hand expert takes as input the entire frame and performs hand detection as well as pose estimation. It detects hands in 81.5 \% of the images, which means  that an average of 18.5 \% of the hand poses are missing for each video. Misdetections are usually due to motion blur (Fig.~\ref{fig:posebert_DexYCB_2}) or heavy occlusion of the hand by the manipulated object (Fig.~\ref{fig:posebert_DexYCB_1}). Plugging \posebert on top of this method solves most of these issues and reduces the hand pose error by a large margin.
We observe a significant gain in all metrics, for instance the MPJPE (resp. PA-MPJPE) decreases by 36.92 \% (resp. 57 \%).

\begin{table*}
	\centering
	\resizebox{\textwidth}{!}{
\begin{tabular}{ccllll}
\toprule
 Detection & Regression & \multicolumn{1}{c}{MPJPE $\downarrow$} & \multicolumn{1}{c}{PA-MPJPE $\downarrow$} & \multicolumn{1}{c}{Accel $\downarrow$}
 \\
\midrule 
\color{red}{$\times$} & HR-Net \cite{chao:cvpr2021} & 17.34 & 6.83 & 12.77 \\
\color{red}{\bf{(Ground-truth)}}& ~~~~+ \bf{\posebert}  & \bf{14.05} \errorgain{3.29} & \bf{4.09} \errorgain{2.79} & \bf{3.62} \errorgain{9.15} \\
\midrule 
\color{green}{\checkmark} & LCR-Net - Hand expert ~\cite{lcrnet++} &  46.31 & 16.15 & 33.44 \\
& \textit{(matched groundtruths only)} & \textit{34.51} & \textit{10.07} & \textit{39.81} \\
& ~~~~~~~~~~~~~~~~~~~~~~~~~ \textit{(miss detections replaced by nearest detection)} &  \textit{40.73} & \textit{11.10} & \textit{27.43} \\
& + \bf{\posebert} & \bf{29.21} \errorgain{17.1} & \bf{6.88} \errorgain{9.27} & \bf{4.52} \errorgain{28.92} \\ 
\bottomrule
\end{tabular}
}
\caption{\textbf{State-of-the-art results on DexYCB.} We add \posebert on top of existing image-based methods either by assuming ground-truth detections or by performing detection at inference time. Results are reported in millimeters.}
\label{tab:posebert_impact_hand}
\end{table*}

\subsubsection{Ablation study for the training strategy}
\label{sub:ablation_hand}
We study the impact on training when varying the length of the input sequence, the percentage of time-steps replaced by random pose, and the level of noise added to the pose.
We also compare against a simple and robust median filtering baseline.
By default, we train \posebert without any masking or noise perturbation, and with a sequence of length 76. Results are reported in Table \ref{tab:posebert_ablation_bis_DexYCB}.

We observe that increasing the sequence length leads to better performance on all metrics.
This indicates that taking into account a large enough temporal window is important for updating the initial pose with the surrounding ones.
We study the impact of random masking and observe that increasing the masking ratio at training time always increases the performance at inference time.
Injecting gaussian noise in the input leads to a boost in performances as long as the noise is big enough, and particularly impacts the Accel metric which measures the smoothness of the pose sequence.
Replacing a percentage of the poses by random ones (taken from the batch) also bring an improvement to all metrics and impact the Accel most.
Finally, we mix all these training strategy and observe that they are all complementary.

\begin{table*}
	\centering
	\resizebox{0.7\textwidth}{!}{
\begin{tabular}{ll|ccc|ccc}
\toprule
 & & \multicolumn{6}{c}{\textit{Image-based model}} \\
 & & \multicolumn{3}{c}{HR-Net \cite{chao:cvpr2021}} & \multicolumn{3}{c}{LCR-Net - Hand expert \cite{lcrnet++}} \\
 \cmidrule(lr){3-8} 
 \multicolumn{2}{l|}{\textit{Video-based model}} & MPJPE & PA-MPJPE & Accel & MPJPE & PA-MPJPE & Accel \\
\midrule
& Baseline & 17.34 & 6.83 & 12.77 & 46.31 & 16.15 & 33.44 \\ 
& ~~~ + Median filtering & 16.46 & 6.69 & 4.31 & \textit{36.58} & \textit{11.31} & \textit{5.89} \\
\midrule
& ~~~ + \bf{\posebert} & \\
(a) & ~~~~~~~~~~ Seq. length ~~~~~~~~ 76        &  \bf{16.95} & \bf{6.12} & \bf{7.96} & \bf{52.06} & \bf{11.35} & \bf{38.00} \\
(b) & ~~~~~~~~~~~~~~~~~~~~~~~~~~~~~~~~~~~~  32  &  16.99 & 6.68 & 8.98 & 52.59 & 11.89 & 39.65  \\
(c) & ~~~~~~~~~~~~~~~~~~~~~~~~~~~~~~~~~~~~~~  8 &  17.12 & 6.75 & 11.1 & 53.11 & 12.54 & 39.44   \\
\midrule
(d) & ~~~~~~~~~~ Masking ~~~~~~~~~  10\%      & 16.74 & 5.99 & \bf{7.62} & 40.69 & 9.29 & 27.49 \\
(e) & ~~~~~~~~~~~~~~~~~~~~~~~~~~~~~~~~~~ 25\% & 16.73 & 5.99 & 8.18 & 39.75 & 9.34 & 26.90 \\
(f) & ~~~~~~~~~~~~~~~~~~~~~~~~~~~~~~~~~~ 50\% & \bf{16.51} & \bf{5.79} & 8.30 & \bf{38.91} & \bf{8.94} & \bf{26.32} \\
\midrule 
(g) & ~~~~~~~~~~ Noise joints ~~~~~ 1e-2      & \bf{15.92} & 3.29 & \bf{5.12} & \bf{38.92} & \bf{6.32} & \bf{18.58}\\
(h) & ~~~~~~~~~~~~~~~~~~~~~~~~~~~~~~~~~~ 1e-3  & 16.09 & \bf{2.96} & 6.75 & 42.30 & 6.60 & 28.56 \\
(i) & ~~~~~~~~~~~~~~~~~~~~~~~~~~~~~~~~~~ 1e-4  & 16.78 & 4.40 & 9.75 & 50.23 & 8.74 & 40.21 \\
\midrule
(j) & ~~~~~~~~~~ Random poses ~~10\%           & 16.21 & 6.25 & 8.72 & 42.54 & 10.15& 18.97\\
(k) & ~~~~~~~~~~~~~~~~~~~~~~~~~~~~~~~~~~ 25\%  & 16.05 & 5.97 & 6.83 & \bf{39.30} & 9.03 & 14.96 \\
(l) & ~~~~~~~~~~~~~~~~~~~~~~~~~~~~~~~~~~ 50\%  & \bf{15.29} & \bf{4.57} & \bf{4.59} & 40.23 & \bf{8.42} & \bf{9.33} \\
\midrule
& ~~~~~~~~~~ Mix (a)+(e)+(g)+(h)+(k)  & \bf{14.04} & \bf{4.09} & \bf{3.62} & \bf{29.21} & \bf{6.88} & \bf{4.52} \\
\bottomrule
\end{tabular}
}
\caption{\textbf{Additional ablation on the \posebert hyperparameters on DexYCB (split s0).}
We study the impact of different perturbations at training time. The default \posebert does not have any masking, random poses or noise, and is trained with sequences of length 76.
}
\label{tab:posebert_ablation_bis_DexYCB}
\end{table*}


\subsubsection{Robustness to miss detections}
\label{sub:miss_detections}

\begin{figure}
    \centering
    \includegraphics[width=\linewidth]{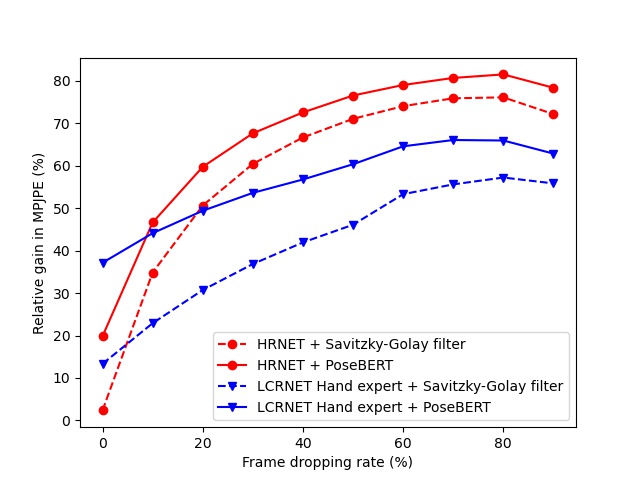}  
    \caption{\textbf{Impact of frame dropping on the relative gain of \posebert against the image-based model.} We report the relative gain on the MPJPE reported in millimeters on DexYCB test set.}
    \label{fig:frame_dropping}
\end{figure}

To make sure that \posebert is robust against missing poses, we perform an analysis on frame dropping reported in Figure~\ref{fig:frame_dropping}.
At test time, we manually drop a percentage (from 0 to 90\%) of the frames of the input pose sequence.
We compare \posebert against a smoothing baseline, again the Savitzky-Golay filter~\cite{Savitzky-Golay}.
We report the relative gain in MPJPE against the image-based model.

First, we run this analysis using HRNET as image-based model.
Since this image-based model is using the ground-truth 2D hand bounding boxes as input there is no missing detections.
Moreover, HRNET is trained on the associated training images so there is not a domain gap and the estimated poses are not very noisy.
We observe that \posebert always bring a better relative gain to the image-based model compared to the smoothing baseline.
The relative gain of \posebert ranges from 20 \% for 0 frames dropped to 80\% with 90\% of missing input poses.

Second, we run the same analysis taking the poses estimated by the LCR-Net Hand expert as input.
We observe a similar conclusion; \posebert brings a significantly better relative gain compared to the smoothing baseline.
However the gap between \posebert and the Savitzky-Golay is larger when using this single image based model instead of HRNET.
This can be explained by the fact that LCR-Net Hand expert is not trained on the DexYCB dataset and is also detecting hands in the entire image.
Hence, the input sequences fed to the temporal module are likely to be quite noisy and \posebert plays an important role at filtering and producing robust hand meshes.

\subsubsection{Future frame prediction}
\label{sub:hand_future}

\begin{figure}
\centering
\small
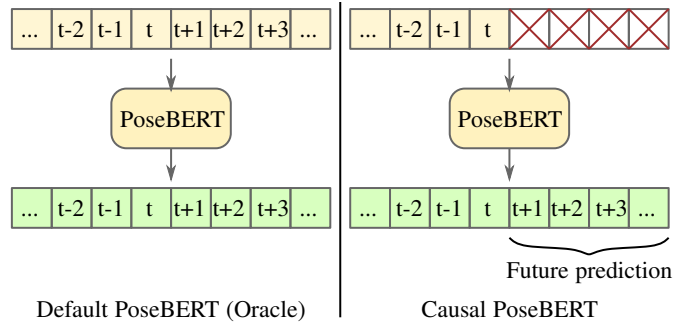
\caption{\label{fig:future_prediction_schematics}\textbf{Future prediction using PoseBERT}. Left: PoseBERT takes as input a sequence of observations and predicts a denoised sequence of poses. Right: By padding a sequence of observations with a mask token, one can predict future poses.}
\end{figure}

We benchmark \posebert on the task of future hand pose prediction.
Given 15 observed frames (0.5 second), we leverage \posebert to predict the next 30 frames (see Figure~\ref{fig:future_prediction_schematics}).
We separate the analysis of the results on the following time horizons: 5, 10, 15 and 30 frames in the future.
For a fair comparison, we provide 3 baselines: \textit{No-velocity} corresponds to predicting the last estimated hand pose into the future, \textit{Velocity propagation} means that we compute the angular velocity between the two last observed frames and predict the future poses by propagating this velocity in the next frames. Finally, \textit{Oracle} is an upper bound which observes the sequence up to the timestep of interest and runs \posebert. Results are reported in Table~\ref{tab:future_predictions_DexYCB}.

We observe that \posebert always outperforms the \textit{No-velocity} and \textit{Velocity propagation} baselines for all horizons, and across all image-based models except for LCR-Net Hand expert at horizon $\Delta_t = 5$. The \textit{Velocity propagation} baseline achieves good performances at horizon $\Delta_t = 5$, however it quickly deteriorates and become worse than the  \textit{No-velocity} baseline. Indeed, a constant movement cannot persist for too long or the pose becomes unrealistic. How long should a movement last and which movement should come next is the kind of prior that is very hard to hand-craft and that we learn with \posebert.
We conclude that even if \posebert is not trained to predict future pose, it can be repurposed successfully for this task.

\begin{table}[h]
\centering
\resizebox{\linewidth}{!}{
\begin{tabular}{ll|cc|cc}
\toprule
 && \multicolumn{4}{c}{\textit{Image-based model}} \\
 && \multicolumn{2}{c}{HR-Net \cite{chao:cvpr2021}} & \multicolumn{2}{c}{LCR-Net - Hand expert \cite{lcrnet++}} \\
 \cmidrule(lr){3-6}
 Horizon & Method & MPJPE $\downarrow$ & PA-MPJPE $\downarrow$ & MPJPE $\downarrow$ & PA-MPJPE $\downarrow$ \\
\midrule
\multirow{4}{*}{$\Delta t=5$} & No-velocity & 28.72 & 7
16 & 45.82 & 8.59 \\ 
& Velocity Propagation    &  24.50 & 6.53 & \bf{38.51} & 9.30\\
& \bf{\posebert}   &  \bf{24.44} & \bf{6.41} & 44.41 & \bf{8.13} \\ 
& \textit{Oracle}  & \textit{18.10} & \textit{5.51} & \textit{36.40} & \textit{7.71} \\ 
\midrule
\multirow{4}{*}{$\Delta t=10$} & No-velocity & 36.47 & 8.14 & 51.33 & 9.09 \\ 
& Velocity Propagation    & 47.84 & 10.96 & 58.78 & 13.42 \\
& \bf{\posebert}   &  \bf{28.29} & \bf{6.64} & \bf{50.34} & \bf{8.38} \\ 
& \textit{Oracle}  & \textit{17.35} & \textit{5.30} & \textit{35.55} & \textit{7.81} \\
\midrule
\multirow{4}{*}{$\Delta t=15$} & No-velocity & 45.47 & 9.02 & 57.47 & 9.66\\ 
& Velocity Propagation    &  72.83 & 15.24 & 81.04 & 17.32 \\
& \bf{\posebert}   &  \bf{32.87} & \bf{6.93} & \bf{55.64} & \bf{8.72} \\ 
& \textit{Oracle}  & \textit{16.89} & \textit{5.43} & \textit{35.38} & \textit{8.00} \\
\midrule
\multirow{4}{*}{$\Delta t=30$} & No-velocity & 61.57 & 9.95 & 69.01 & 10.38 \\ 
& Velocity Propagation    & 110.89 & 22.87 & 117.01 & 24.13 \\
& \bf{\posebert}   &  \bf{35.97} & \bf{6.99} & \bf{61.64} & \bf{9.32} \\ 
& \textit{Oracle}  & \textit{14.67} & \textit{4.71} & \textit{30.70} & \textit{7.82} \\
\bottomrule
\end{tabular}
}
\caption{\textbf{Future frames prediction on DexYCB}.
Given 15 observed frames we predict the future hand pose from 5 to 30 frames into the future.
\textit{No-velocity} corresponds to predicting the last estimated hand pose into the future.
\textit{Velocity propagation} means that we compute the future hand motion using the last observed velocity (\ie $p_{t+1}:=p_{t}+\delta_t $) and propagate it frame by frame in the future.
\textit{Oracle} corresponds to the results we can achieve if we observe the sequence up to the frame we want to predict.
Results are reported in millimeters.
}
\label{tab:future_predictions_DexYCB}
\end{table}

\subsubsection{Application example: robotic teleoperation}
\label{sub:hand_application}

\begin{figure}
\centering
\includegraphics[width=\linewidth]{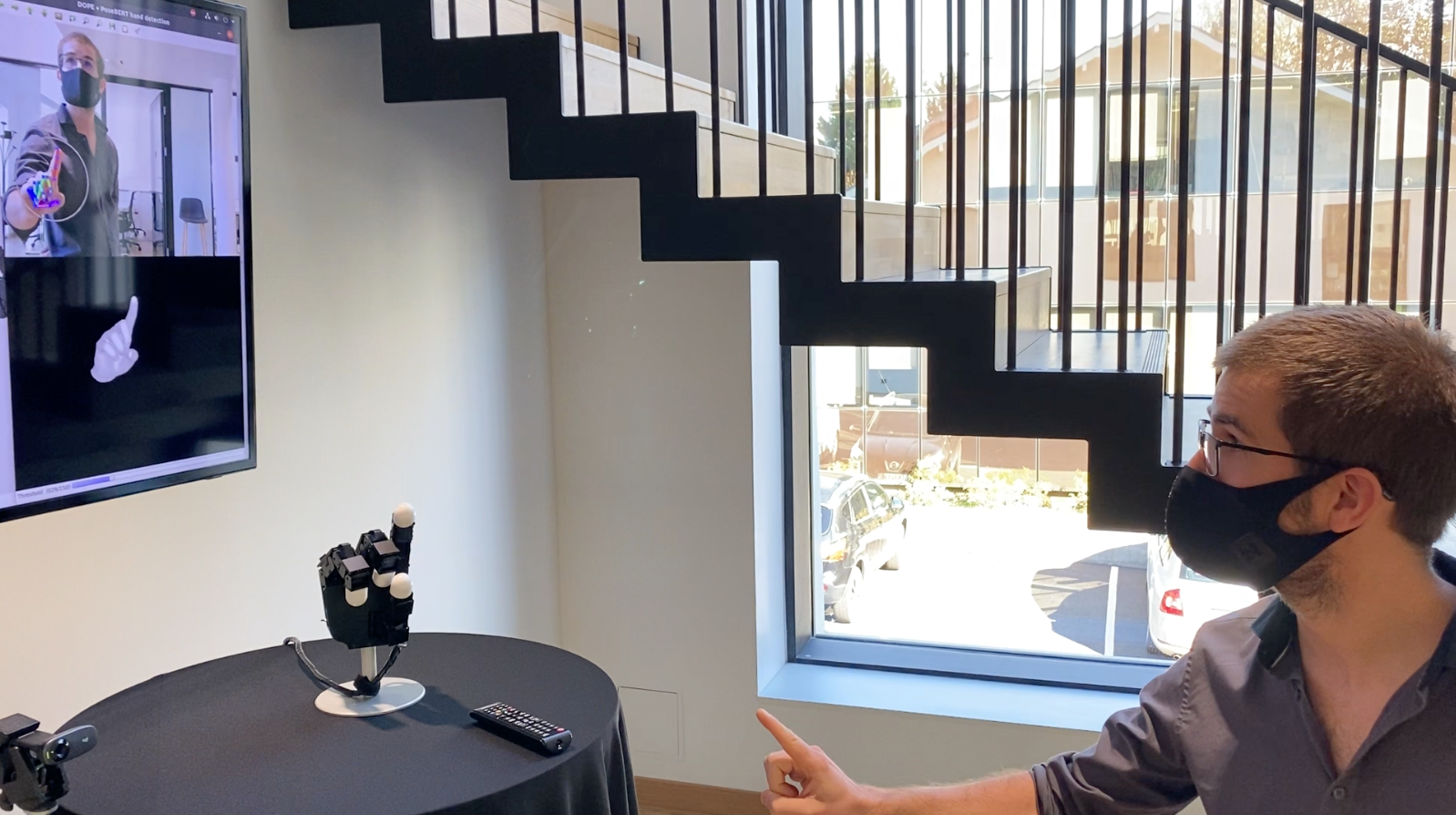} \\
\caption{\label{fig:robot_teleoperation}
\textbf{Application example: using PoseBERT to animate a robotic gripper using an RGB webcam.}
A human hand is captured using a webcam (bottom left corner). Hand pose is estimated using LCR-Net hand expert and fed into \posebert for obtaining a smooth and robust estimation of MANO, in real-time. We use kinematic retargeting to transfer this pose to the one of the robot gripper.
A video of this demo and other qualitative examples are available at \url{https://europe.naverlabs.com/blog/posebert/}.
}
\end{figure}

Being able to predict the pose of a human in real-time enables a large variety of practical applications -- for example robotic teleoperation. To demonstrate such a use case, we developed an application in which a person remotely animates a robotic gripper using his/her right hand.
Figure~\ref{fig:robot_teleoperation} shows our actual setup.
Using LCR-Net++ \cite{lcrnet++},
we detect hands appearing in images of a webcam RGB stream, and we predict the location of their 2D and 3D keypoints. Data is processed at 30Hz using a \emph{Nvidia RTX 5000} GPU.
At each timestep, we process detections of the last 64 frames using PoseBERT and produce a corresponding sequence of MANO~\cite{romero2017mano} hand parameters. We use the last pose of the sequence as our predicted hand pose.
We convert this hand pose into a target pose for our robotic gripper -- an Allegro hand, from Wonik Robotics -- using a kinematic retargeting procedure inspired by DexPilot~\cite{handa2020dexpilot}. The robotic hand continuously follow such target pose using a simple servoing loop.

Using PoseBERT in such application provides two advantages.
First, the denoising and smoothing capabilities of PoseBERT are useful to filter and smooth the target commands.
Second, latency can be a critical problem for teleoperation. We therefore experimented using PoseBERT to predict future hand poses as a way to reduce this latency. We qualitatively observed reasonably good hand predictions up to 15 frames in the future, leading to a more reactive system.


\section{Limitations of \posebert}
\label{sub:limittaions}

\begin{figure*}[th]
\centering
\setlength{\synthfigwidth}{\linewidth}
\begin{subfigure}{\textwidth}
\includegraphics[width=\synthfigwidth]{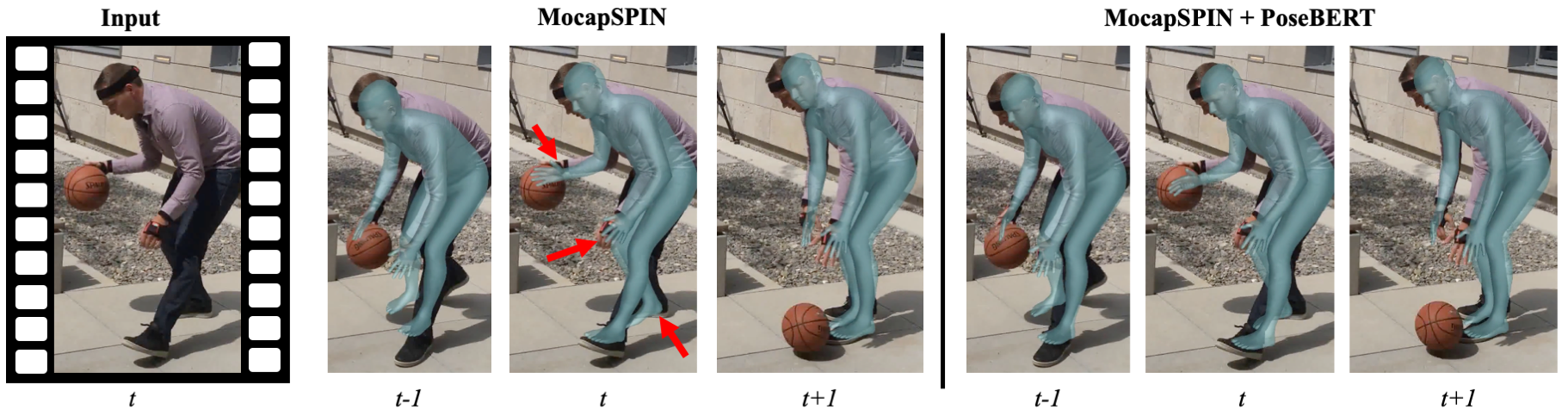}
\caption{\label{fig:xxxxx} \textit{Failure case 1.}
The image-based method (MocapSPIN) is better predicting the 3D locations of hands and feet compared to \posebert for time step $t$, as shown by the red arrows.
MocapSPIN is performing worse than \posebert on the previous and next time steps however ($t-1$ and $t+1$).
\vspace{0.2cm}}
\end{subfigure}
\begin{subfigure}{\textwidth}
\setlength{\synthfigwidth}{\linewidth}
\includegraphics[width=\synthfigwidth]{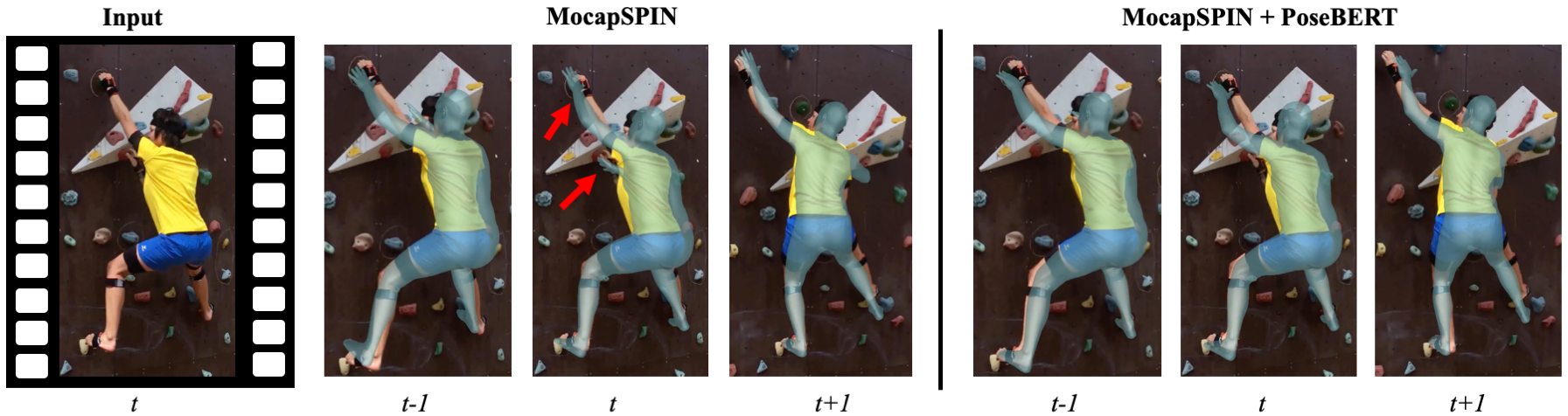}
\caption{\label{fig:xxxx}
\textit{Failure case 2.}
MocapSPIN is correctly predicting the position of the right hand at time $t$ even though it is almost fully occluded, as shown by the red arrow.
However MocapSPIN is producing low quality predictions for the surrounding time steps $t-1$ and $t+1$.
These erroneous predictions negatively impact the final estimation of \posebert at time step $t$.
}
\end{subfigure}
\caption{\label{fig:failure}
\textbf{Failure cases of \posebert.}
We provide qualitative examples of common failure cases, where initial 3D mesh estimates are better than those obtained after applying \posebert.
}
\end{figure*}



We showed that \posebert can improve the performances on any image-based method for a limited computational overhead.
\posebert is working well using different type of parametric 3D models such as MANO or SMPL.
But we also identified several limitations such as discussed below.


\noindent \textbf{Performance degradation.} We provide in Fig. \ref{fig:failure} some qualitative examples of failure cases where \posebert is degrading the quality of the initial image-based predictions.
Such cases may happen in scenarios of fast human motions such as shown in Fig. \ref{fig:failure}(a) where \posebert is smoothing too much the input 3D poses.
We posit that training \posebert with more diverse motion speeds could help to mitigate this problem.
A similar issue may happen in case of long-term occlusion such as shown in Fig. \ref{fig:failure}(b).
In this example, the image-based model is correctly predicting the human mesh at time $t$ -- especially for the right hand which is almost fully occluded -- but using \posebert degrades the prediction for this time step.
The image-based model is indeed producing low quality predictions for surrounding timestamps $t-1$ and $t+1$, and these predictions provide a wrong temporal context which negatively impacts the output of \posebert.
Such issue could probably be mitigated by adding a notion of confidence in the outputs of the image-based model, or by training \posebert with a noise model more representative of the image-based method used. It would however make \posebert dependent of the image-based model, contrary to the proposed approach which can be applied on top of any image-based pose prediction method.




\noindent \textbf{Deterministic output.} 
\posebert is a fully deterministic neural network producing a unique pose prediction for each time step.
Predicting a pose distribution~\cite{posegpt} instead would allow to handle the uncertainties inherent to monocular pose estimation.  
A variational approach~\cite{VAE,meva} could also be useful if we were to predict human poses more than a few seconds in the future, as such predictions are likely to be highly multimodal.

\noindent \textbf{2D errors.} Finally, we follow standard metrics \cite{spin} for human/body mesh estimation such as MPJPE but those metrics do not take the reprojection of the 3D mesh in the image.
We observe that \posebert, like other mesh estimation methods~\cite{spin,hmmr}, achieves good results in 3D 
but tends to produce inaccurate 2D joints estimations compared to hand/body pose estimations specifically designed for this task.
We therefore believe that future work should explore methods that are good for both 3D and 2D body/hand pose estimation.

\section{Conclusion}
\label{sec:conclusion}

`In this paper we propose a way of leveraging MoCap data to improve video-based human and hand 3D mesh recovery.
We introduce \posebert, a transformer module that directly regresses the parameters of a parametric 3D model from noisy and/or incomplete image-based estimations.
\posebert is purely trained on MoCap data via masked modeling.
Our experiments show that \posebert can be trained for both body or hand 3D model and that can be readily plugged on top of any image-based model to leverage temporal context and improve its performance.

\bibliographystyle{IEEEtran}
\bibliography{main}

\begin{IEEEbiography}[{\includegraphics[width=1in,height=1.25in,clip,keepaspectratio]{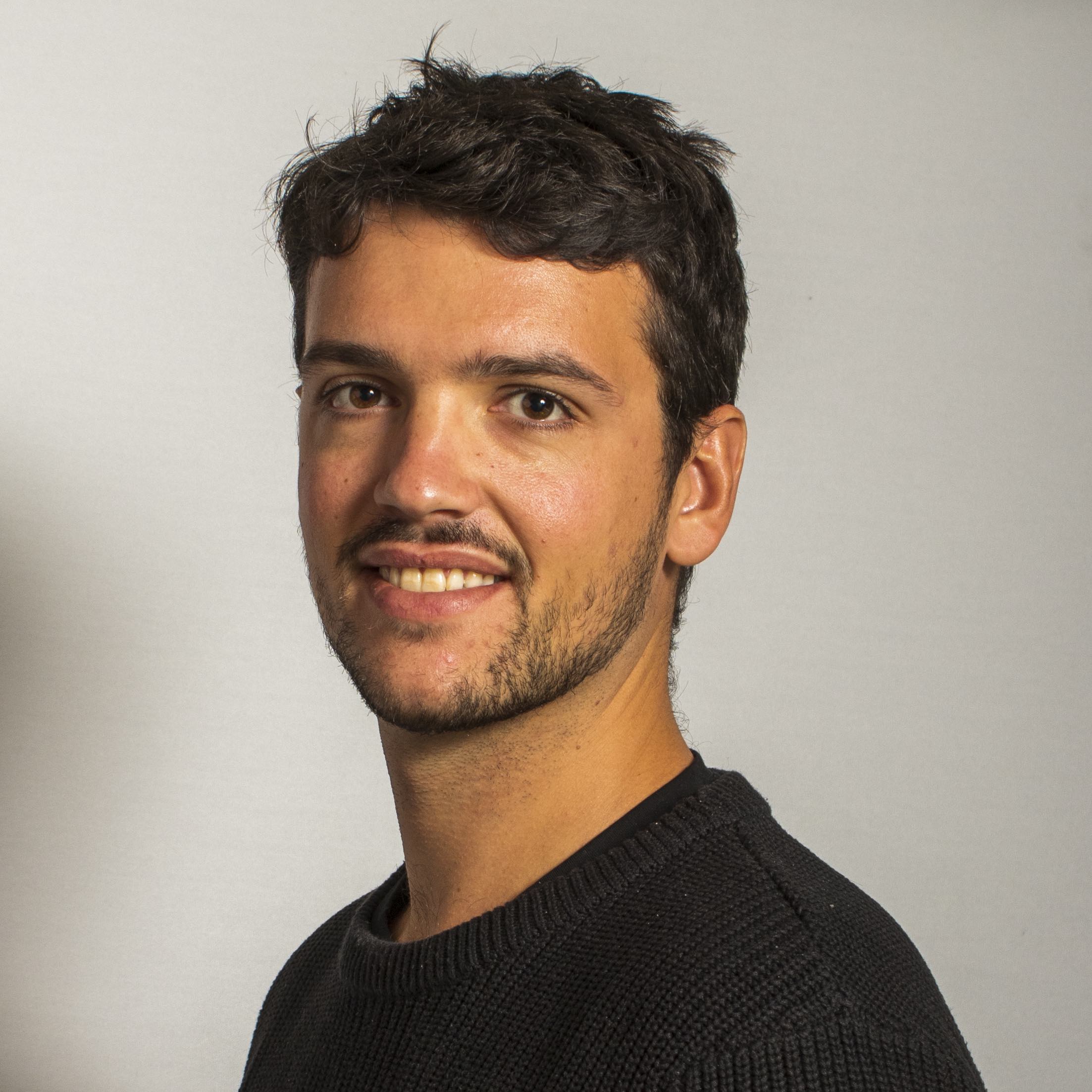}}]{Fabien Baradel} obtained a MsC. degree in Statistics from Ecole Nationale de Statistique and Analyse Information in Rennes, France (2016) and a Ph.D degree in Computer Science from INSA Lyon, France (2020).
He joined NAVER LABS Europe, in Grenoble (France), in 2020 as a Research Scientist where he is working on computer vision and machine learning with. He is focusing on understanding people from visual data.
\end{IEEEbiography}

\begin{IEEEbiography}[{\includegraphics[width=1in,height=1.25in,clip,keepaspectratio]{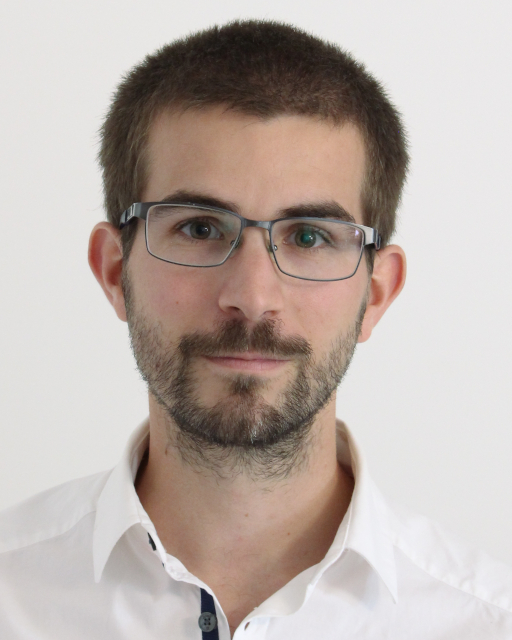}}]{Romain Brégier} obtained a MSc. degree in Engineering from the \'Ecole Centrale de Lyon (2013), a MSc. in Signal and Image Processing from the University of Lyon  (2013), and a Ph.D. in Computer Science in collaboration between Inria Grenoble and Sil\'eane, France (2018). After working at Siléane on machine vision and robotics for industrial applications, he joined NAVER LABS Europe as a researcher in 2019, where he is working on computer vision, geometry and robotics.
\end{IEEEbiography}


\begin{IEEEbiography}[{\includegraphics[width=1in,height=1.25in,clip,keepaspectratio]{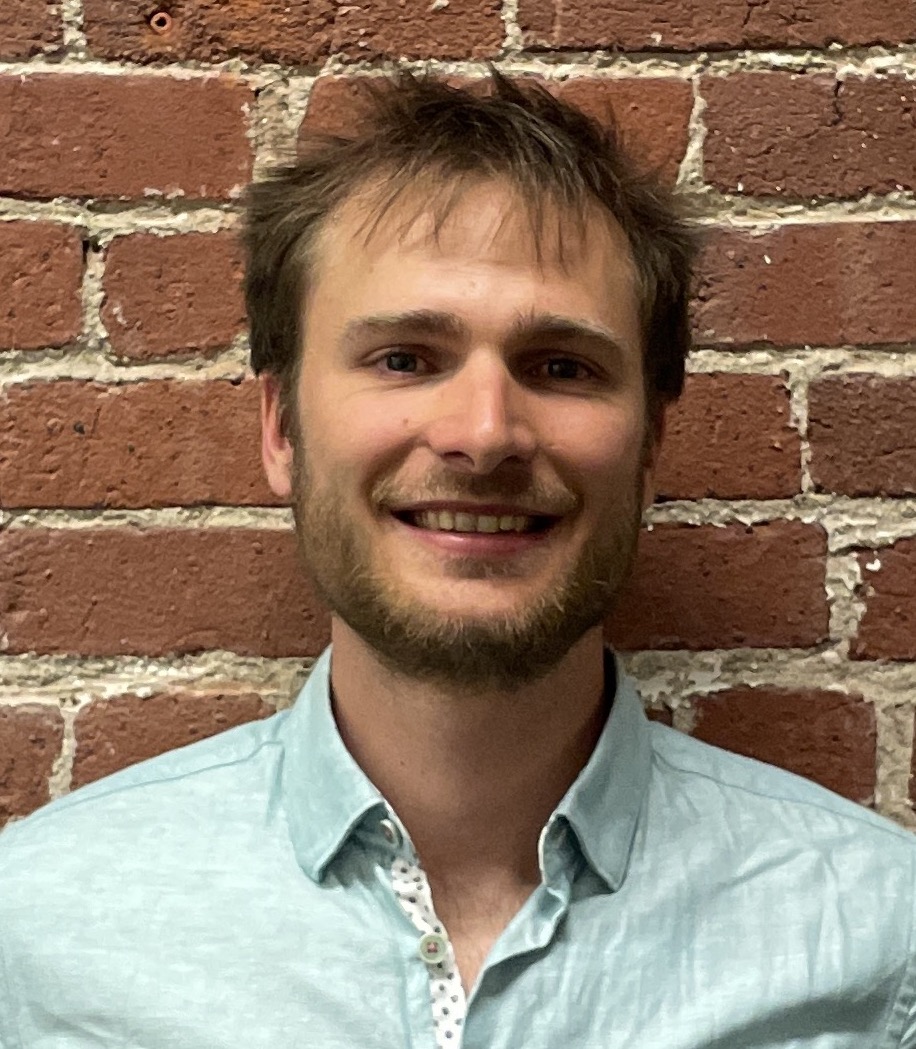}}]{Thibault Groueix} is a research engineer at Adobe Research since April 2021. He received his PhD on Computer Science from École des ponts ParisTech in 2020. He was previously a research scientist at NAVER LABS Europe (2020-2021). He research is centered on 3D Deep Learning, with a particular emphasis on 3D humans.
\end{IEEEbiography}

\begin{IEEEbiography}[{\includegraphics[width=1in,height=1.25in,clip,keepaspectratio]{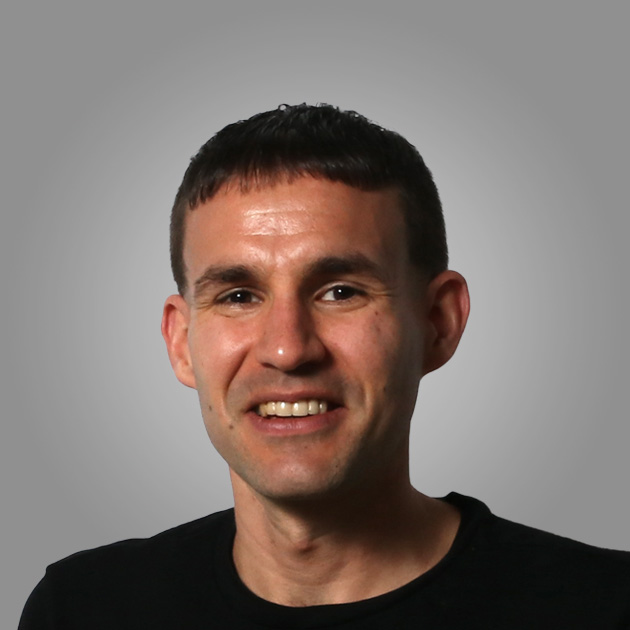}}]{Philippe Weinzaepfel} received a M.Sc. degree from Universite Grenoble Alpes, France, and ´
Ecole Normale Superieure de Cachan, France, ´
in 2012. He was a PhD student in the Thoth
team, at Inria Grenoble and LJK, from 2012 until
2016, and received a PhD degree in computer
science from Universite Grenoble Alpes in 2016. ´
He is currently a Senior Research Scientist at NAVER
LABS Europe, France. His research interests include computer
vision and machine learning, with special interest in representation learning and human pose estimation.
\end{IEEEbiography}

\begin{IEEEbiography}[{\includegraphics[width=1in,height=1.25in,clip,keepaspectratio]{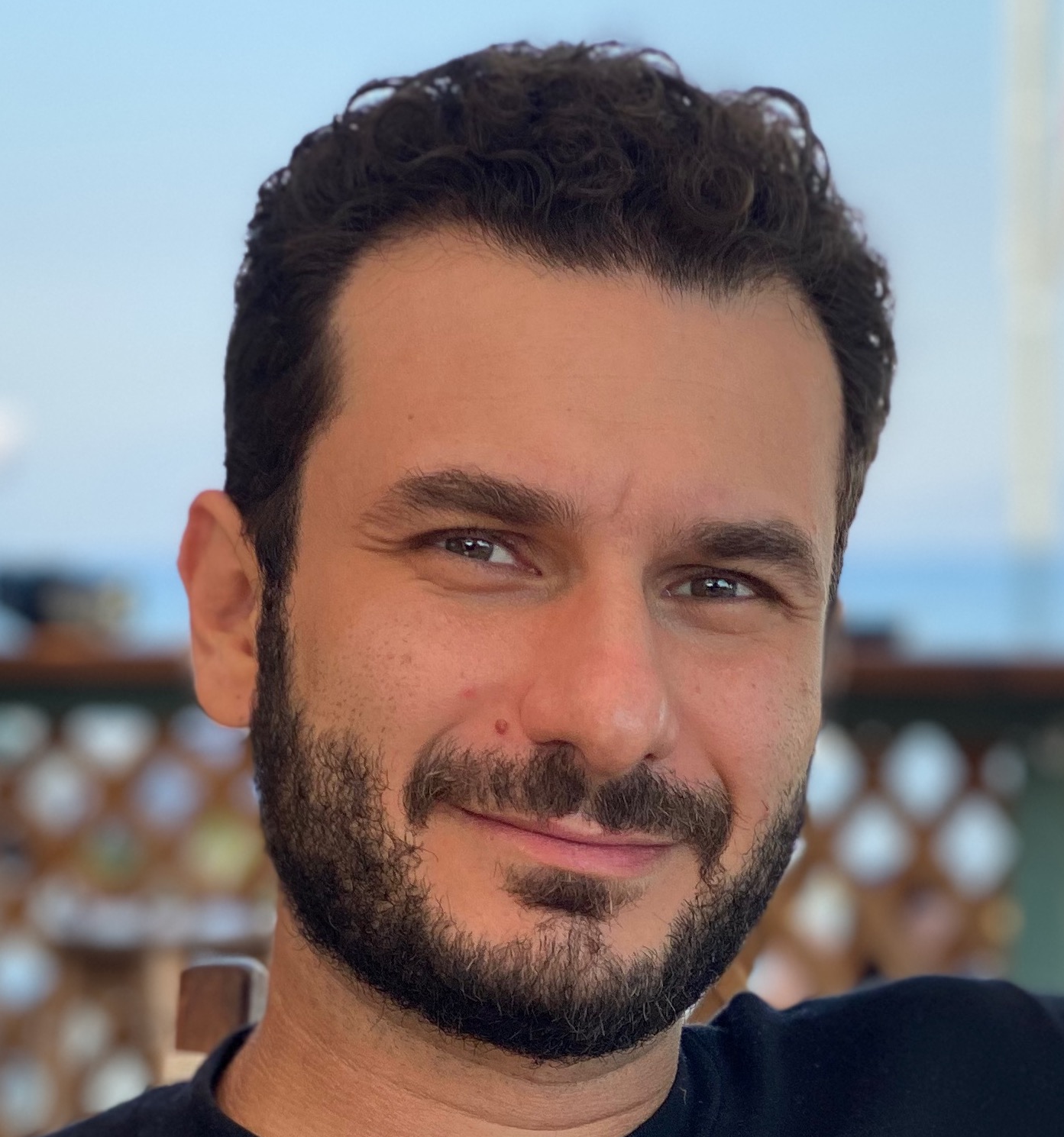}}]{Yannis Kalantidis} is a senior research scientist at NAVER LABS Europe in Grenoble since 2020. He received his PhD on Computer Science from the National Technical University of Athens in 2014. He was a research scientist at Yahoo Research in San Francisco (2015-2017) and a research scientist at Facebook AI in Menlo Park (2017-2019). His research revolves around visual representation and multi-modal learning under limited supervision and resources, as well as adaptive multi-modal systems. He is also passionate about bringing the computer vision community closer to socially impactful tasks, datasets and applications for worldwide impact and co-organized workshops like “Computer Vision for Global Challenges” (CV4GC @ CVPR 2019), “Computer Vision for Agriculture” (CV4A @ ICLR 2020) and “Wikipedia and Multi-Modal \& Multi-Lingual Research” (Wiki-M3L @ ICLR 2022) in top-tier AI venues.
\end{IEEEbiography}

\begin{IEEEbiography}[{\includegraphics[width=1in,height=1.25in,clip,keepaspectratio]{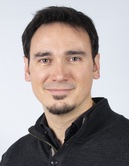}}]{Grégory Rogez} graduated from École Nationale Sup\'erieure de Physique de Marseille (now Centrale Marseille) in 2002 and received the M.Sc. degree in biomedical engineering and the Ph.D. degree in computer vision from the University of Zaragoza, Spain, in 2005 and 2012 respectively. His work on monocular human body pose analysis received the best Ph.D. thesis award from the Spanish Association on Pattern Recognition (AERFAI) for the period 2011-2013. He was a regular visiting research fellow at Oxford Brookes University (2007-2010), a Marie Curie Fellow at the University of California, Irvine (2013-2015),  a Research Scientist with the LEAR/THOTH team at Inria Grenoble Rhône-Alpes (2015-2018) and since 2019 he is a Senior Research Scientist and a Team Lead at NAVER LABS Europe. His research interests include computer vision and machine learning/deep learning, with a special focus on understanding people from visual data. This includes human detection and tracking, 2D/3D human pose estimation, 3D human body shape reconstruction, object manipulation and activity recognition in images and videos.
\end{IEEEbiography}
\end{document}